%% file: main.tex
\documentclass[a4paper, 10pt, conference]{ieeeconf}                                                                \usepackage{FG2026}

\FGfinalcopy

\IEEEoverridecommandlockouts                                                                                                                                                  

\usepackage{upgreek}
\usepackage{graphicx}
\usepackage[table]{xcolor}
\usepackage{subcaption}
\usepackage{pifont}
\usepackage{makecell}
\usepackage{multirow}
\usepackage{booktabs}
\usepackage{cite}
\usepackage{float}
\usepackage{bm}

\usepackage[hidelinks]{hyperref}

\usepackage{amsmath,amssymb}

\def\FGPaperID{228} 
\title{\LARGE \bf
Gaze4HRI: Zero-shot Benchmarking Gaze Estimation Neural-Networks for Human-Robot Interaction
}

\author{\parbox{16cm}{\centering
    {\large Huibert Kwakernaak$^1$ and Pradeep Misra$^2$}\\
    {\normalsize
    $^1$ Faculty of Electrical Engineering, Mathematics and Computer Science, University of Twente, Enschede, The Netherlands\\
    $^2$ Department of Electrical Engineering, Wright State University, Dayton, USA}}
    \thanks{This work was not supported by any organization}}
\author{\parbox{16cm}{\centering
    {\large Berk Sezer, Ali G\"orkem K\"u\c{c}\"uk, Erol \c{S}ahin, and Sinan Kalkan}\\
    {\normalsize
    Dept. of Computer Eng. and ROMER (Robotics Center), Middle East Technical University, Ankara, T\"urkiye\\
        \{berk.sezer, gorkem.kucuk, erol, skalkan\}@metu.edu.tr}
}}

\usepackage{fancyhdr}
\begin{document}

\ifFGfinal
\thispagestyle{empty}
\pagestyle{empty}
\else
\author{Anonymous FG2026 submission\\ Paper ID \FGPaperID \\}
\pagestyle{plain}
\fi
\maketitle
\thispagestyle{fancy} 

\fancyfoot[L]{\scriptsize © 2026 IEEE. Personal use of this material is permitted. Permission from IEEE must be obtained for all other uses, in any current or future media, including reprinting/republishing this material for advertising or promotional purposes, creating new collective works, for resale or redistribution to servers or lists, or reuse of any copyrighted component of this work in other works.}

\begin{figure*}[t]
  \centering
  \includegraphics[width=0.8\textwidth]{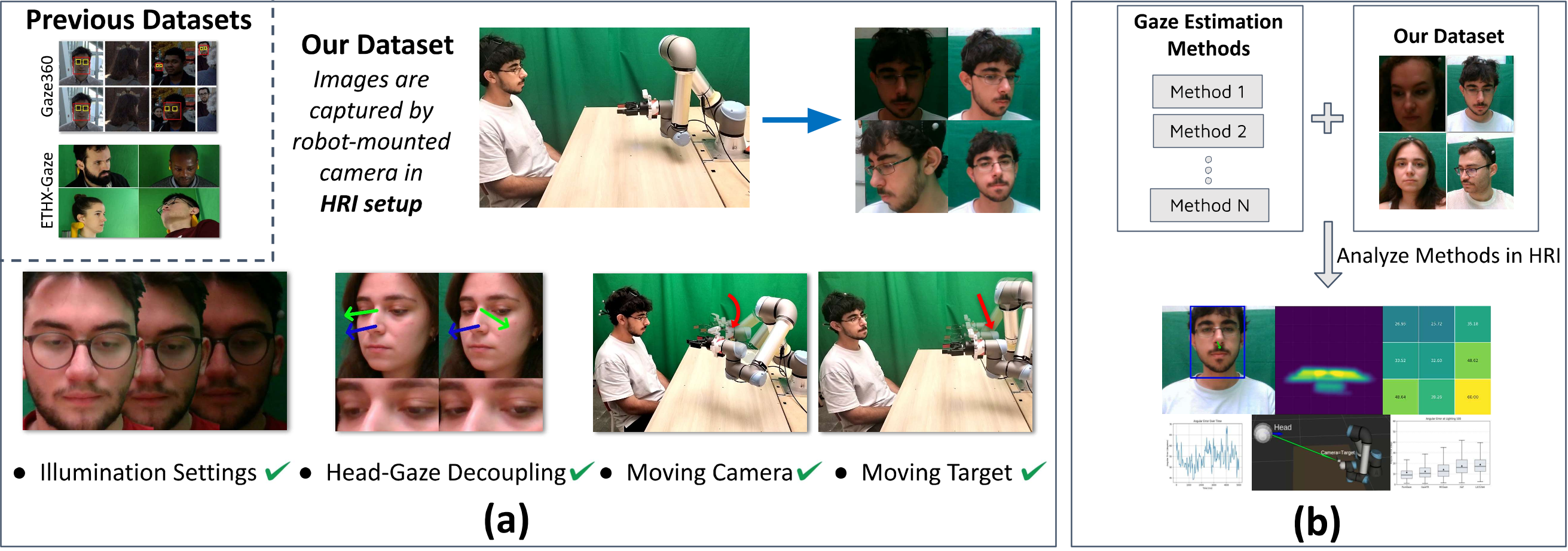}
  \caption{(a) We introduce Gaze4HRI, an extensive benchmark which includes high-quality gaze recordings for 50+ subjects, 3,000+ videos, 600,000+ frames for an HRI setting. (b) Using our dataset, we evaluated the performance of five state-of-the-art gaze estimation architectures for HRI settings.}
  \label{fig:teaser}
\end{figure*}

\begin{abstract}
While zero-shot appearance-based 3D gaze estimation offers significant cost-efficiency by directly mapping RGB images to gaze vectors, its reliability in Human-Robot Interaction (HRI) settings remains uncertain. Existing benchmarks frequently overlook fundamental HRI conditions, such as dynamic camera viewpoints and moving targets in video. Furthermore, current cross-dataset evaluations often suffer from a complexity gap, where methods trained on diverse datasets are tested on significantly smaller and less varied sets, failing to assess true robustness. To bridge these gaps, we introduce \emph{Gaze4HRI}, a large-scale dataset (50+ subjects, 3,000+ videos, 600,000+ frames) designed to evaluate state-of-the-art performance against critical HRI variables: illumination, head-gaze conflict, as well as the motion of camera and gaze target in video. Our benchmark reveals that all evaluated methods fail in at least one condition, identifying steeply-downward gaze as a universal failure point. Notably, \emph{PureGaze} trained on the \emph{ETH-X-Gaze} dataset uniquely maintains resilience across all other conditions. These results challenge the recent focus in the literature on complex spatial-temporal modeling and Transformer-based architectures. Instead, our findings suggest that extensive data diversity, as exemplified by the \emph{ETH-X-Gaze} dataset, serves as the primary driver of zero-shot robustness in unconstrained environments, while resilience-enhancing frameworks, such as \emph{PureGaze's} self-adversarial loss for gaze feature purification, provide a substantial further improvement. Ultimately, this study establishes a rigorous benchmark that provides practical guidelines for practitioners as well as reshaping future research. The dataset and codes are available at \url{https://gazeforhri.github.io}.
\end{abstract}


\section{Introduction}
Gaze provides a continuous stream of attentional and social information that shapes the quality of human interaction, with applications spanning driver safety~\cite{sharma2024review}, assistive robotics~\cite{FischerJanzen2024}, Industry 5.0~\cite{hri_industry_5.0}, and AR/VR~\cite{Lin2025} (see \cite{ghosh2022automatic, pathirana2022survey, cheng2024appearance} for reviews). 

While gaze estimation can utilize specialized hardware such as infrared cameras or depth sensors using feature- or model-based techniques that fit explicit geometric eye models \cite{gaze_importance_survey, eye_model_rgbd_camera}, appearance-based methods that directly regress gaze from standard RGB images~\cite{cheng2024appearance} are becoming increasingly prominent, due to their simplicity (not requiring specialized sensors) and scalability (ability to track multiple subjects within a single image). Especially, the learning-based approach towards developing appearance-based gaze estimation methods by training them on large datasets is showing much promise. In particular, extra data needed to fine-tune pre-trained methods requires the availability of ground-truth gaze vector information, which involves special eye-tracker or motion-capture systems; consequently, most practitioners prefer zero-shot deployment of appearance-based methods instead.

\textbf{Gaze in Human-Robot Interaction (HRI).} Estimation of gaze is crucial for HRI behaviors, such as mutual gaze (gazing at each other), shared gaze (gazing at the same point in space)
~\cite{Schreiter2024THORMAGNI, Schreiter2024HeadGaze, jording2018social, AdmoniScassellati2017, kompatsiari2017importance, bayesian_gaze_estimation_hri_intro, nonverbal_hri_nod_gaze}, to enable robots initiate and maintain engagement with humans. 
Although gaze estimation methods have been developed and evaluated in general settings, their suitability and zero-shot performance in HRI settings remain uncertain.

\textbf{Research Gaps.} To our knowledge, no study quantitatively evaluates zero-shot state-of-the-art (SOTA) performance in HRI settings where the camera is attached to a moving robot engaging with shared and mutual gazing. Moreover, understanding how key environmental variables affect performance is essential for users to estimate method accuracy for their use case. In terms of environmental variables, illumination and pose variation have been extensively explored as key challenges in the literature \cite{Zhang2020ETHXGaze, kellnhofer2019gaze360, cheng2024appearance}. Beyond these, gaze-specific factors such as head-gaze conflict are reported \cite{guan2023mcgaze}, yet remain under-explored (which this work uniquely explores). Most importantly, the accuracy of SOTA gaze methods under HRI-induced conditions, such as dynamic targets and ego-motion from wrist-mounted cameras, has yet to be established in the literature.

\textbf{Contributions.} We address the aforementioned research gaps with the following contributions (see also Fig. \ref{fig:teaser}): 

\begin{itemize}
    \item We introduce \emph{Gaze4HRI}, a large-scale benchmark (50+ subjects, 600,000+ frames) simulating object-centered and mutual gaze tasks of HRI. 
    
    \item Through \emph{Gaze4HRI}, we systematically analyze the performance of SOTA methods by dissecting how variations in illumination, pose, head-gaze conflict, and movement affect gaze estimation errors.

\end{itemize}

\section{Related Work}

\subsection{Appearance-Based Datasets}
Appearance-based datasets are summarized in \autoref{tab:dataset_comparison}~\cite{cheng2024appearance, Zhang2020ETHXGaze}.\footnote{Additional dataset features are tabulated in the Supp. Mat. Also, Tab. \ref{tab:dataset_comparison} writes ``Synthetic Diversity'' for \emph{UT Multiview}, since Sugano et al. add illumination variation during image reconstruction, while using a single illumination condition during data collection \cite{sugano2014utmultiview}.} While earlier datasets \cite{fischer2018rtgene, krafka2016gazecapture, zhang2017mpiigaze, sugano2014utmultiview, funes2014eyediap} were fundamental to the field, the most prominent datasets today for unconstrained settings are \emph{ETH-X-Gaze}~\cite{Zhang2020ETHXGaze} and \emph{Gaze360}~\cite{kellnhofer2019gaze360}. \emph{ETH-X-Gaze} provides 1M+ images under extreme poses and controlled illumination, while \emph{Gaze360} offers diverse \emph{in-the-wild} video comprising 200+ subjects (though it suffers some label noise \cite{kellnhofer2019gaze360, Zhang2020ETHXGaze}). Current trends of cross-dataset evaluation involve training on these high-complexity sets and testing on smaller, less diverse datasets from prior work such as \emph{MPIIGaze} and \emph{EYEDIAP}~\cite{cheng2022puregaze, cheng2022gazetr, vuillecard2025gat}, providing an incomplete assessment of robustness. \emph{Gaze4HRI} bridges this complexity gap by benchmarking in a complex HRI setting.

\subsection{Appearance-Based Methods} Appearance-based methods have progressed from localized eye-patch processing~\cite{zhang2017mpiigaze, sugano2014utmultiview} to context-rich full-face inputs~\cite{krafka2016gazecapture, Zhang2020ETHXGaze}, culminating in SOTA Transformer and spatio-temporal architectures~\cite{cheng2022gazetr, guan2023mcgaze, efe, vuillecard2025gat}\footnote{There also exists domain adaptation methods like ADDA~\cite{Tzeng2017ADDA}, GVBGD~\cite{Cui2020GVBGD}, UMA~\cite{Cai2020UMA}, DAGEN~\cite{Guo2020DAGEN}, and PNP-GA~\cite{Liu2021PNPGA} that improve cross-domain accuracy, but they require samples from the target domain for alignment, and therefore are not zero-shot.}.

\textbf{Evaluated Methods.} To benchmark zero-shot performance; we selected five prominent, appearance-based neural-networks that: (i) Did 3D-Gaze estimation. (ii) Were released after 2020. (iii) Had open-source, pretrained weights. Thus, we selected: \emph{PureGaze}~\cite{cheng2022puregaze}, \emph{GazeTR}~\cite{cheng2022gazetr}, \emph{L2CS-Net}~\cite{abdelrahman2022l2cs}, \emph{MCGaze}~\cite{guan2023mcgaze}, and \emph{GaT}~\cite{vuillecard2025gat}. All utilize ResNet for feature extraction. \emph{L2CS-Net} and \emph{PureGaze} are CNN-based; the former regresses pitch \& yaw separately, while the latter uses a self-adversarial framework to purify features of identity and illumination bias~\cite{cheng2022puregaze}. Conversely, \emph{GazeTR}, \emph{MCGaze}, and \emph{GaT} leverage Transformers to capture global dependencies, with the latter two serving as temporal models that incorporate spatial-temporal context from short video clips.

\textbf{Training Dataset of Evaluated Methods.} The selected methods utilize different training datasets in their publicly-released versions: \emph{PureGaze} and \emph{GazeTR} are trained on \emph{ETH-X-Gaze}, while \emph{L2CS-Net}, \emph{MCGaze}, and \emph{GaT} are trained on \emph{Gaze360}. While training on heterogeneous datasets precludes a direct architectural comparison, the primary objective of this study is to evaluate the overall \emph{method}—which consists of the inseparable union of architecture and training data for a learning-based method. This focus reflects the practical reality for end-users prioritizing end-to-end performance over architectural design.

In addition to the pre-trained weights, we trained \emph{PureGaze} and \emph{GazeTR} on the \emph{Gaze360} dataset to test Cheng et al.’s claim that \emph{ETH-X-Gaze} offers superior domain generalization for these methods \cite{cheng2022puregaze, cheng2022gazetr}. Specifically, we investigate whether this assertion holds in the challenging context of HRI
\footnote{Conversely, training video-based methods on \emph{ETH-X-Gaze} contradicts their design, as it provides static frames rather than videos. Thus, neither the original authors nor we trained these models on it \cite{guan2023mcgaze, vuillecard2025gat}.}.

\newcommand{\modcell}[2]{\makecell[l]{#1 {\scriptsize #2}}}

\begin{table*}[t]
\caption{Comparison of Appearance-Based Gaze Datasets. Key attributes: Number of subjects and images, \emph{Setting} specifies the setting used during data collection, \emph{Illumination (Controlled/Diversity)} distinguishes between controlled vs. uncontrolled, as well as specifying diversity (Discrete Conditions, Synthetic, or Natural Diversity); \emph{Head Pose/Gaze Range} values are reported as (Pitch, Yaw); and \emph{Movement} specifies the source of temporal variation across consecutive video frames.}
\label{tab:dataset_comparison}
\begin{center}
\footnotesize
\setlength{\tabcolsep}{3.5pt} 
\renewcommand{\arraystretch}{2.0}
\begin{tabular}{|l||c|c|c|c|c|c|c|}
\hline
\textbf{Dataset} & \textbf{\# Subjects} & \textbf{\# Images} & \textbf{\makecell{Setting}} & \textbf{\makecell{Illumination\\(Controlled/Diversity)}} & \textbf{\makecell{Head Pose \\ Range}} & \textbf{\makecell{Gaze \\ Range}} & \textbf{\makecell{Movement\\(For Video)}} \\
\hline
\modcell{EYEDIAP}{(2014) \cite{funes2014eyediap}} & 16 & 426K & Lab. & \makecell{Cont. /\\1 Condition} & $\pm30^{\circ}, \pm15^{\circ}$ & $\pm20^{\circ}, \pm25^{\circ}$ & Moving Target \\
\hline
\modcell{UT Multiview}{(2014) \cite{sugano2014utmultiview}} & 50 & 1.1M & Lab. & \makecell{Cont. /\\Synthetic Div.} & $\pm36^{\circ}, \pm36^{\circ}$ & $\pm36^{\circ}, \pm50^{\circ}$ & N/A (Not Video) \\
\hline
\modcell{MPIIGaze}{(2015) \cite{zhang2017mpiigaze}} & 15 & 213K & Laptop & \makecell{Uncont. /\\Natural Div.} & $\pm30^{\circ}, \pm15^{\circ}$ & $\pm20^{\circ}, \pm20^{\circ}$ & N/A (Not Video) \\
\hline
\modcell{GazeCapture}{(2016) \cite{krafka2016gazecapture}} & 1,474 & 2.45M & Mobile & \makecell{Uncont. /\\Natural Div.} & $\pm40^{\circ}, \pm30^{\circ}$ & $\pm20^{\circ}, \pm20^{\circ}$ & N/A (Not Video) \\
\hline
\modcell{RT-GENE}{(2018) \cite{fischer2018rtgene}} & 15 & 122K & \makecell{Unconstrained \\ Indoor} & \makecell{Uncont. /\\Natural Div.} & $\pm40^{\circ}, \pm40^{\circ}$ & $\pm40^{\circ}, \pm40^{\circ}$ & Moving Camera \\
\hline
\modcell{Gaze360}{(2019) \cite{kellnhofer2019gaze360}} & 238 & 172K & \makecell{Unconstrained \\ In/Outdoor} & \makecell{Uncont. /\\Natural Div.} & N/A, $\pm90^{\circ}$ & $-50^{\circ}, \pm140^{\circ}$ & Moving Target \\
\hline
\modcell{ETH-X-Gaze}{(2020) \cite{Zhang2020ETHXGaze}} & 110 & 1.08M & Lab. & \makecell{Cont. /\\16 Conditions} & $\pm80^{\circ}, \pm80^{\circ}$ & $\pm70^{\circ}, \pm120^{\circ}$ & N/A (Not Video) \\
\hline
\hline
\textbf{\modcell{Gaze4HRI}{(2025) [Ours]}} & 52 & 620K & HRI (Lab.) & \makecell{Cont. /\\4 Conditions} & $\pm75^{\circ}, \pm85^{\circ}$ & \makecell{$-70/+20^{\circ}$,\\$\pm100^{\circ}$} & \makecell{Moving Camera\\\& Target} \\
\hline
\end{tabular}
\end{center}
\end{table*}

\textbf{Input-Output of Evaluated Methods.} 
All evaluated methods share a common input--output structure. On the input side, methods operate either on single frames (\textit{PureGaze}, \textit{GazeTR}, \textit{L2CS-Net}) or on short temporal clips (\textit{MCGaze}: 7 frames, \textit{GaT}: 8 frames). These inputs are pre-processed (cropped or rectified) face regions rather than full frames. Since each method can be deployed with its best input configuration, we analyze the best configuration of each method in subsequent experiments of this study (given in Tab. \ref{tab:method_architecture}), and this is discussed in detail in the Supp. Mat. For output, each method predicts a gaze direction in 3D space, represented either as a vector in Cartesian coordinates $(x,y,z) \in \mathbb{R}^3$ or spherical angles (\textit{pitch}, \textit{yaw}), which  are mathematically equivalent.

\begin{table}[t]
\caption{Configurations of Evaluated Methods.}
\label{tab:method_architecture}
\begin{center}
\footnotesize
\setlength{\tabcolsep}{4pt}
\renewcommand{\arraystretch}{1.2}
\begin{tabular}{|l||c|c|c|}
\hline
\textbf{Method} & \textbf{Pre-processing} & \textbf{Input Type} & \textbf{Resolution} \\
\hline
PureGaze (2022) \cite{cheng2022puregaze} & Rectified & Single Image & $224 \times 224$ \\
\hline
GazeTR (2022) \cite{cheng2022gazetr}  & Rectified & Single Image & $224 \times 224$ \\
\hline
L2CS-Net (2022) \cite{abdelrahman2022l2cs} & Cropped             & Single Image & $448 \times 448$ \\
\hline
MCGaze (2023) \cite{guan2023mcgaze}   & Cropped             & 7-Frame Video & $448 \times 448$ \\
\hline
GaT (2025) \cite{vuillecard2025gat}      & Cropped             & 8-Frame Video & $224 \times 224$ \\
\hline
\end{tabular}
\end{center}
\vspace*{-0.2cm}
\end{table}

\section{Gaze4HRI Dataset}

\subsection{Data Collection Setup}

\begin{figure}[t]
  \centering
  \includegraphics[width=0.7\linewidth]{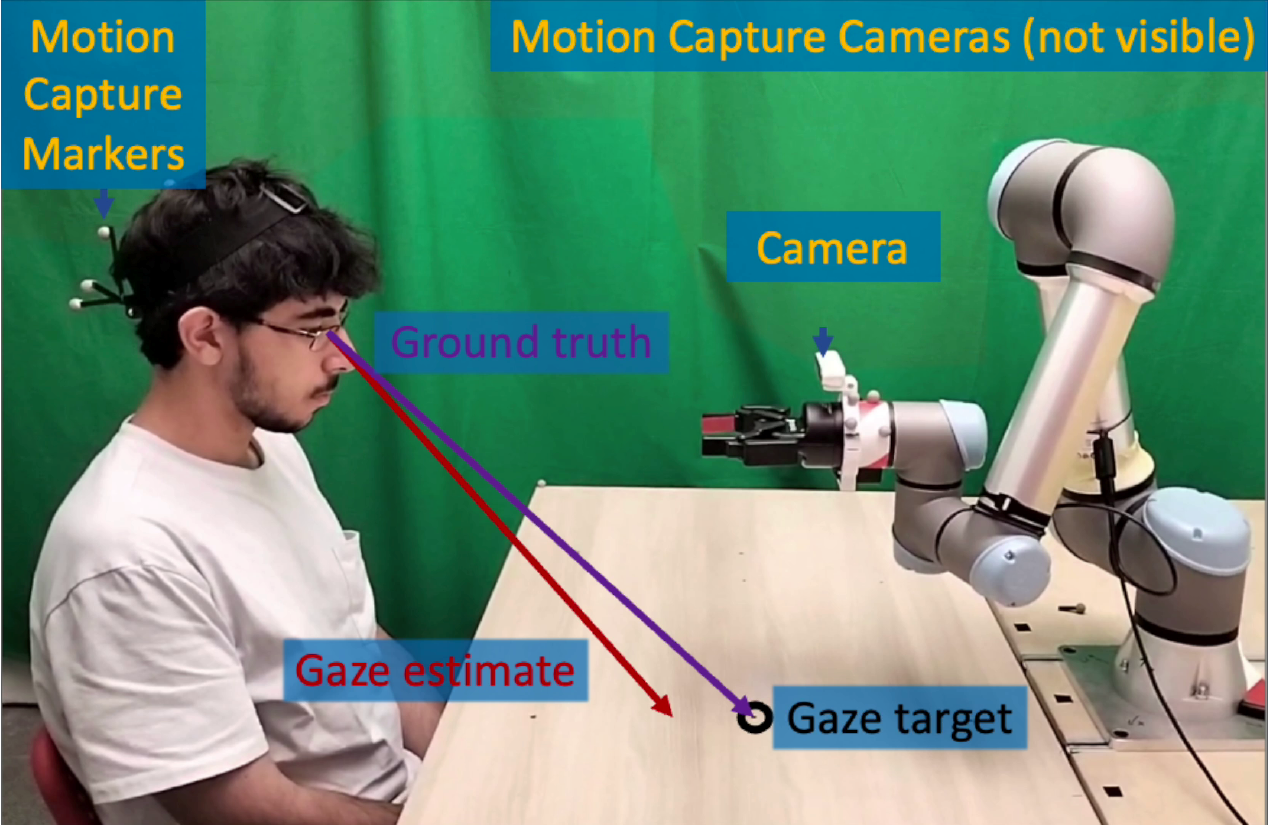}
  \caption{Experimental Setup. The user is instructed to look at a gaze target (a point on the table in the object-centered setup). The ground truth for the gaze is computed through tracking the MoCap markers on the headband and the gaze target (see text). The image from the RGB camera, fixated on the robot, is fed to a deep-learning method, and the resulting gaze estimate is used to evaluate its performance.}
  \label{fig:gaze_ground_truth}\label{fig:setup}
\end{figure}

We designed a setup (Fig. \ref{fig:setup}) centered around a UR5 robot arm (Universal Robots, Denmark). An Intel RealSense D435i camera is attached to the wrist-link of the robot. The environment contains an OptiTrack motion capture system (OptiTrack Inc., USA) to track the head pose of the subject via a trackable headband. This setup enabled the streaming of motion capture data at 100 Hz, and  RGB images from the camera at 30 Hz in \(1920\times1080\) resolution.\footnote{The camera is rigidly attached to the end-effector of the robot, as shown in Fig.~\ref{fig:setup}, and its relative pose with respect to the end-effector of the robot is measured. This allowed us to control and track the pose of the camera in the world through the use of the forward kinematics model and precise sensors of the UR5 robot arm.}

\begin{figure}[t]
  \centering
  \begin{minipage}[t]{0.49\linewidth}
    \centering
    \includegraphics[width=0.7\linewidth]{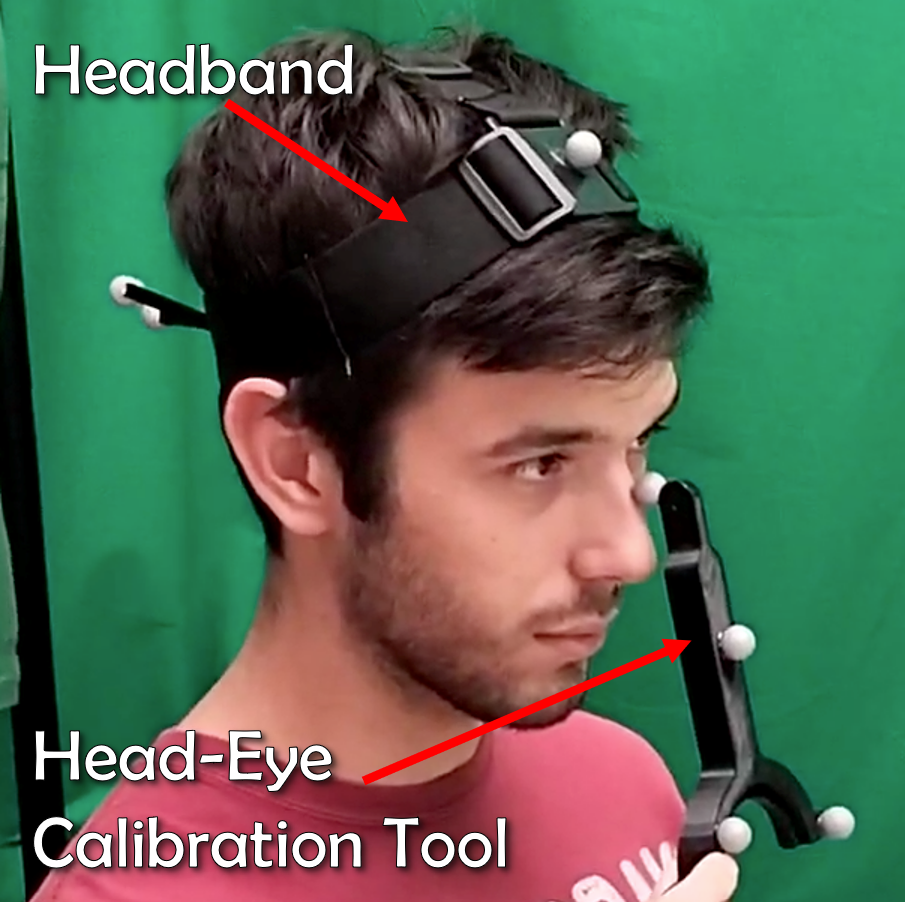}\\
    \textbf{\footnotesize (a) Calibration before experiments}
  \end{minipage}
  \begin{minipage}[t]{0.49\linewidth}
    \centering
    \includegraphics[width=0.7\linewidth]{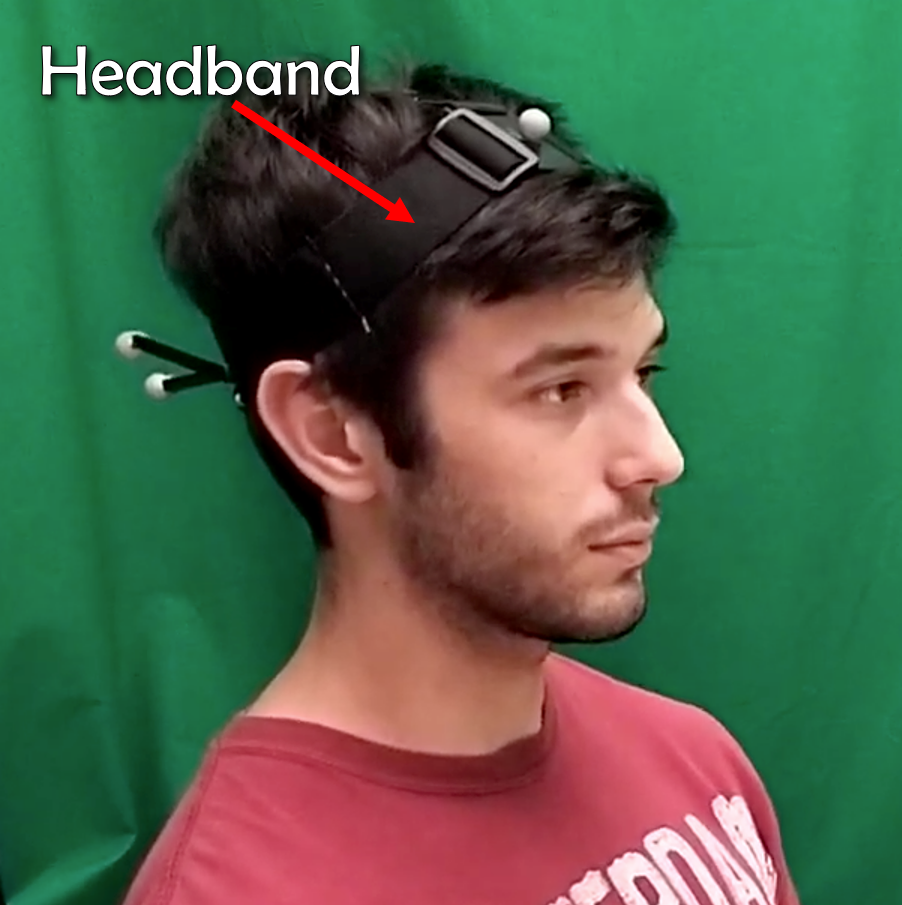}\\
    \textbf{\footnotesize (b) During experiments.}
  \end{minipage}
  \caption{Eye-head calibration for gaze ground truth collection: (a) Head-to-eye transformation is obtained with the help of motion-capture markers before experiments. (b) During experiments, only the head marker is tracked, and the eyes' positions are calculated with the head-eye transformation.}
  \label{fig:head_eye_calib}
\end{figure}

Following prior work \cite{cheng2024appearance}, our ground-truth gaze is defined as the translation vector from the interpupillary midpoint to the gaze target, in the camera frame. To calculate this, we combine the head and eye poses tracked via the motion-capture system with the known environment gaze targets. Specifically, we employ a pre-experiment head-eye calibration procedure by placing a trackable tool above the subject's interpupillary midpoint to measure interpupillary midpoint's pose relative to the trackable headband (Fig.~\ref{fig:head_eye_calib}a). During experiments  (Fig.~\ref{fig:head_eye_calib}b), the motion-capture system tracks the headband, and the midpoint position is automatically inferred to determine the ground-truth vector.\footnote{The pose of the interpupillary midpoint relative to the head is a static transformation, i.e., it does not change during the experiment and it is sufficient to measure it once before the experiment.} Notably, while \emph{Gaze360} relies on the AlphaPose neural network for tracking~\cite{kellnhofer2019gaze360}, our OptiTrack-based approach is substantially more accurate, achieving a tested accuracy of $\pm 0.5 \text{ mm}$ -- see the Supp. Mat. for more details. 

\noindent\textbf{Gaze Error.}  
We evaluate method accuracy using the angular error ($\upalpha$) between the ground-truth gaze vector $\mathbf{g} \in \mathbb{R}^3$ (measured with the motion-capture system) and the estimated gaze vector $\hat{\mathbf{g}} \in \mathbb{R}^3$ (predicted by the method) for each frame (see Fig. \ref{fig:gaze_ground_truth}): \begin{equation}
\small
\upalpha = \arccos ( (\mathbf{g} \cdot \hat{\mathbf{g}}) / (\|\mathbf{g}\| \, \|\hat{\mathbf{g}}\|) ).
\label{eq:gaze_error}
\end{equation}
During data recording, each subject was instructed to blink as normal to collect natural data. Blinked frames were later masked for evaluation, as they are invalid for gaze evaluation, consistent with the protocol in \emph{ETH-X-Gaze} \cite{Zhang2020ETHXGaze}.

\subsection{Variables of Interest}
\begin{figure}[hbt!]
  \centering
  \begin{minipage}[t]{0.48\textwidth}
    \centering
    \includegraphics[height=0.39\linewidth]{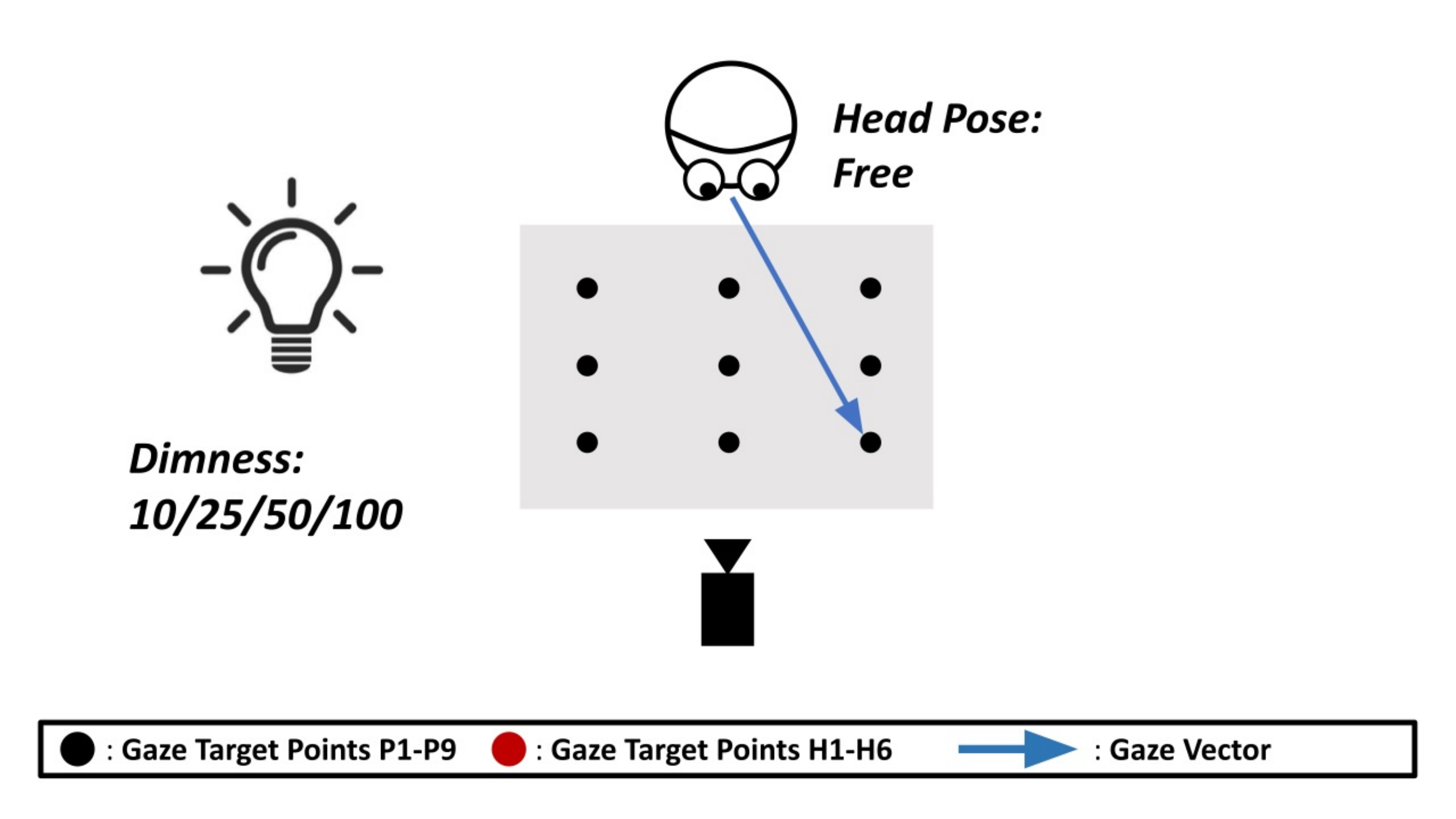}\\
    \textbf{(a) Illumination Setup}
    \phantomsubcaption\label{fig:exp-lighting}
  \end{minipage}\\
  \begin{minipage}[t]{0.48\textwidth}
    \centering
    \includegraphics[height=0.39\linewidth]{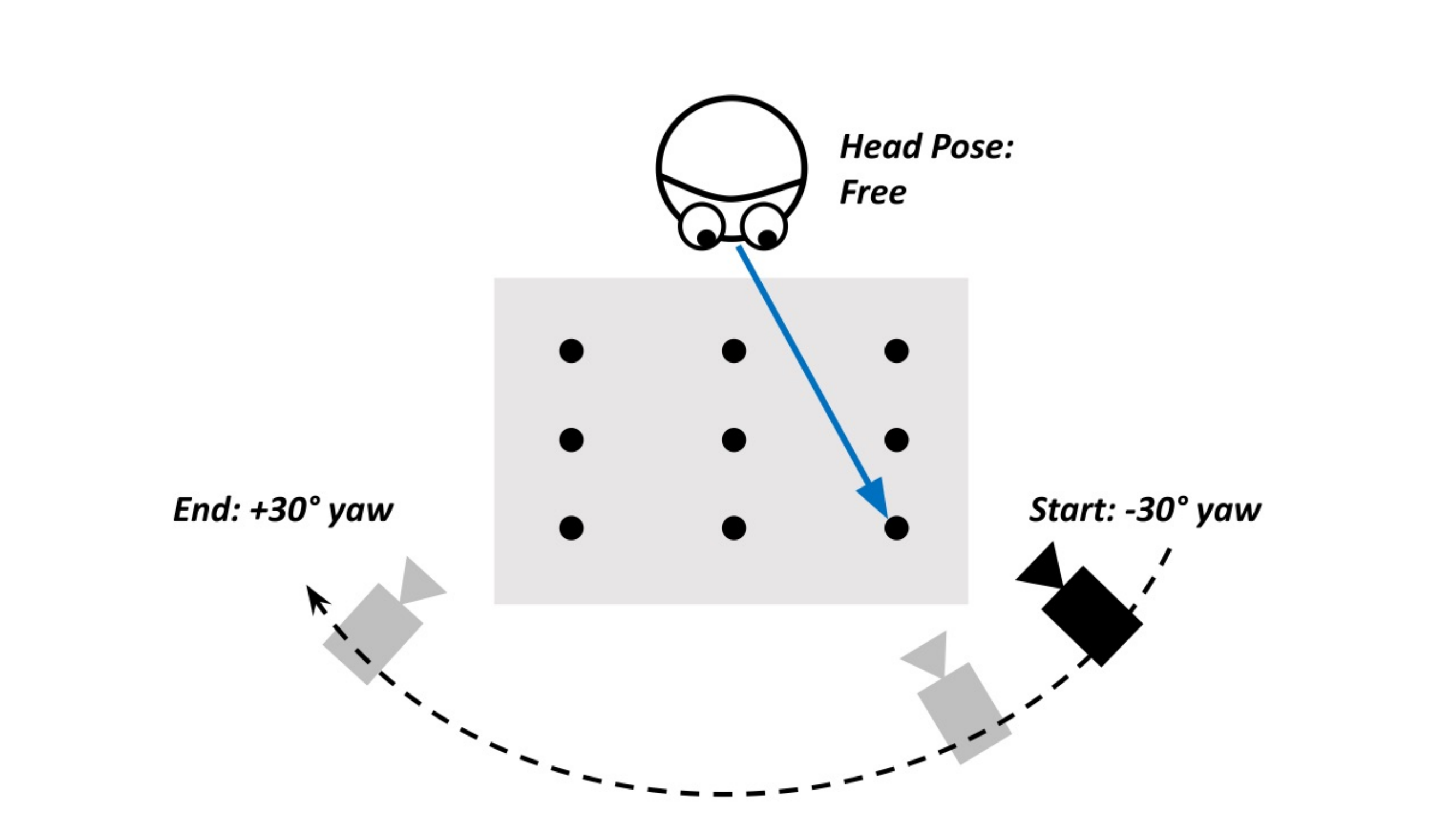}\\
    \textbf{(b) Camera Viewpoint Setup}
    \phantomsubcaption\label{fig:exp-circular}
  \end{minipage}\\
  \begin{minipage}[t]{0.48\textwidth}
    \centering
    \includegraphics[height=0.42\linewidth]{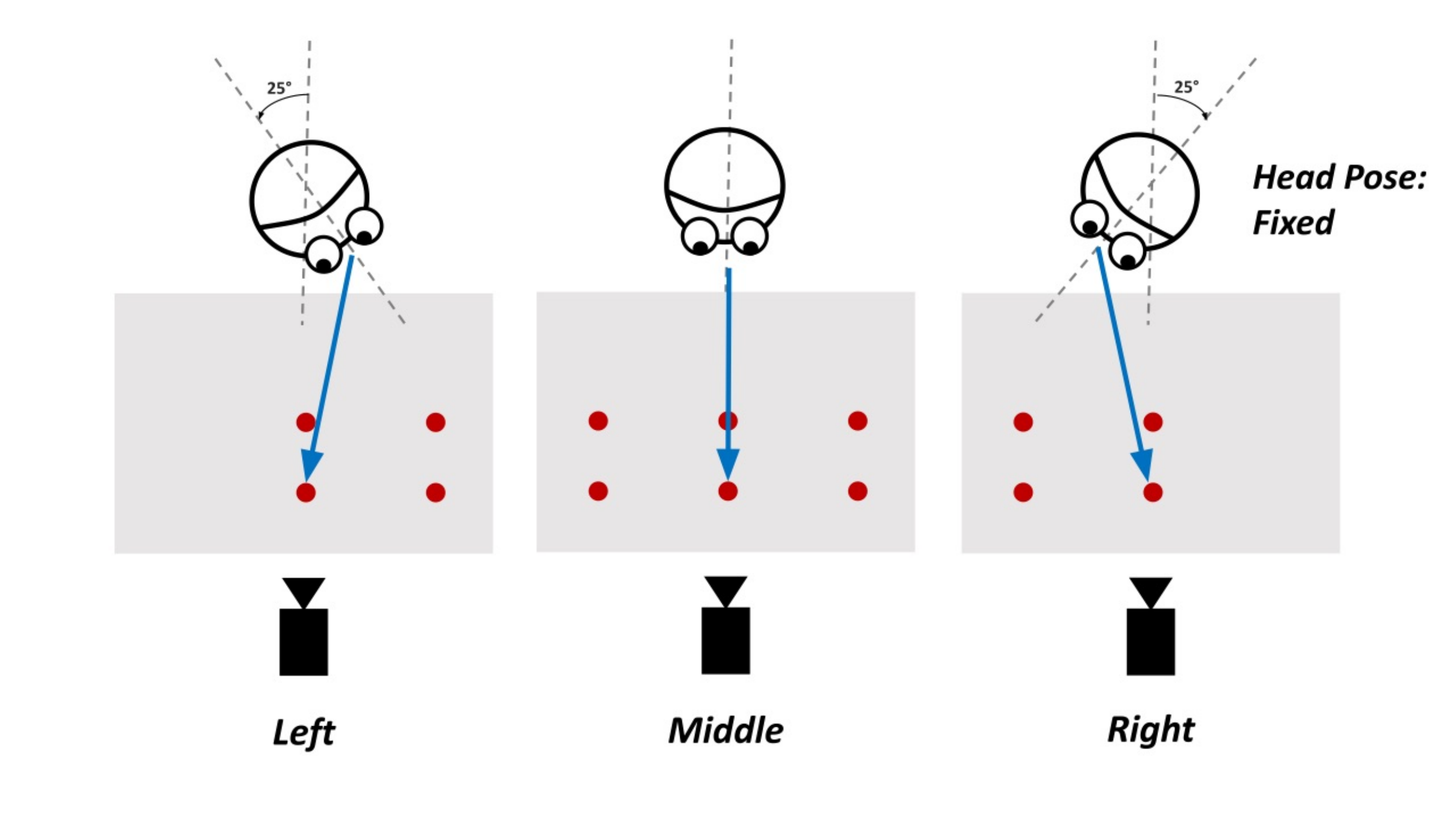}\\
    \textbf{(c) Head-Gaze Conflict Setup}
    \phantomsubcaption\label{fig:exp-headpose}
  \end{minipage}\\
  \begin{minipage}[t]{0.48\textwidth}
    \centering
    \includegraphics[height=0.56\linewidth]{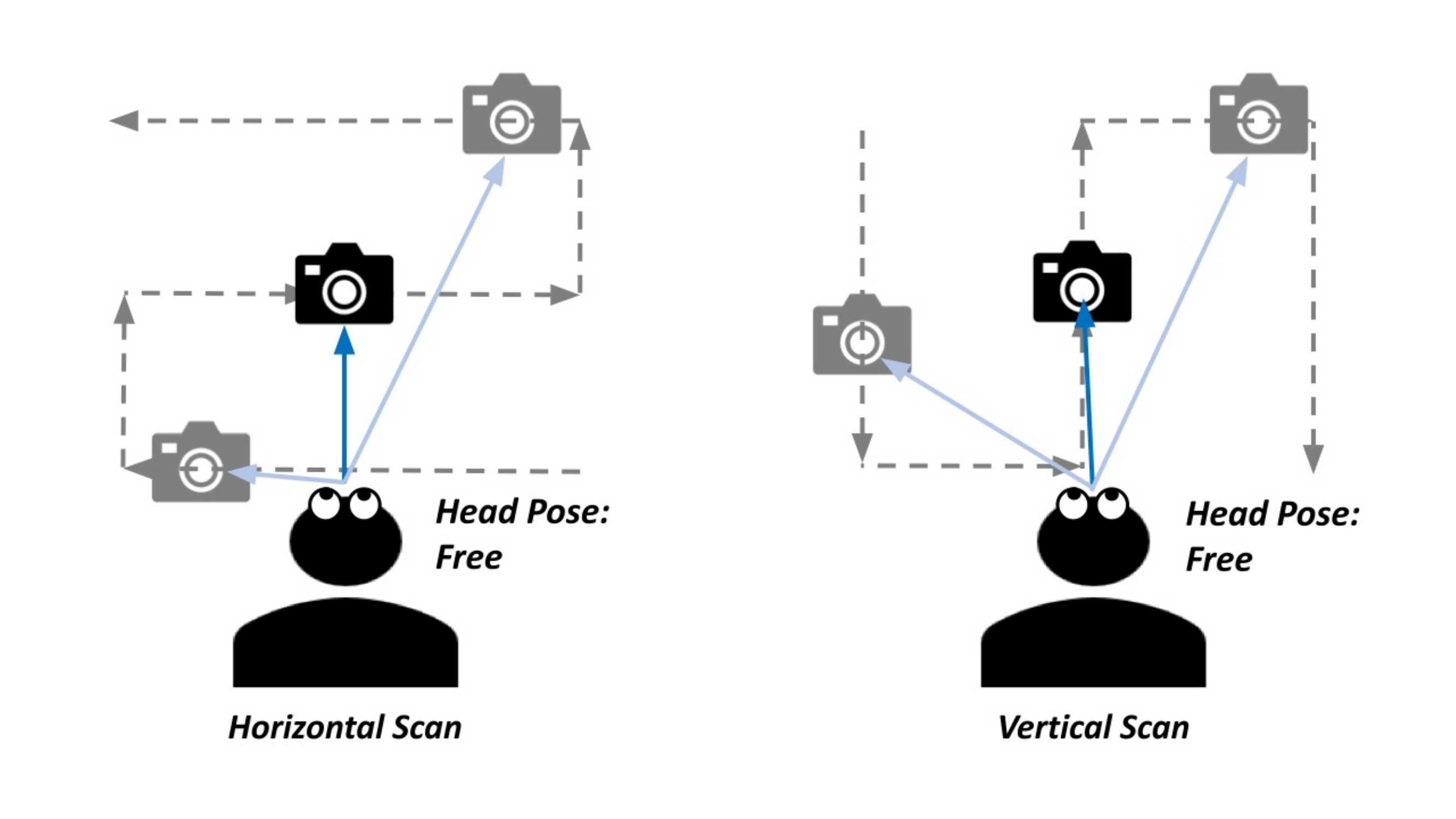}
    \textbf{(d) Moving Target (Mutual Gaze) Setup}
    \phantomsubcaption\label{fig:exp-linear}
  \end{minipage}

  \caption{The four setups used in our analysis.}   \label{fig:experiment_design_all}
\end{figure}

Our setups are designed to systematically test the effect of four  variables: illumination, camera viewpoint, head-gaze conflict, and the gaze target (position variation and movement). \autoref{fig:experiment_design_all} describes the setup to test each variable -- see the respective experiments for details (Sect. \ref{sec:analysis_and_results}).

\subsection{Gaze Targets}
\begin{figure}[hbt!]
  \centering
  \includegraphics[width=\linewidth]{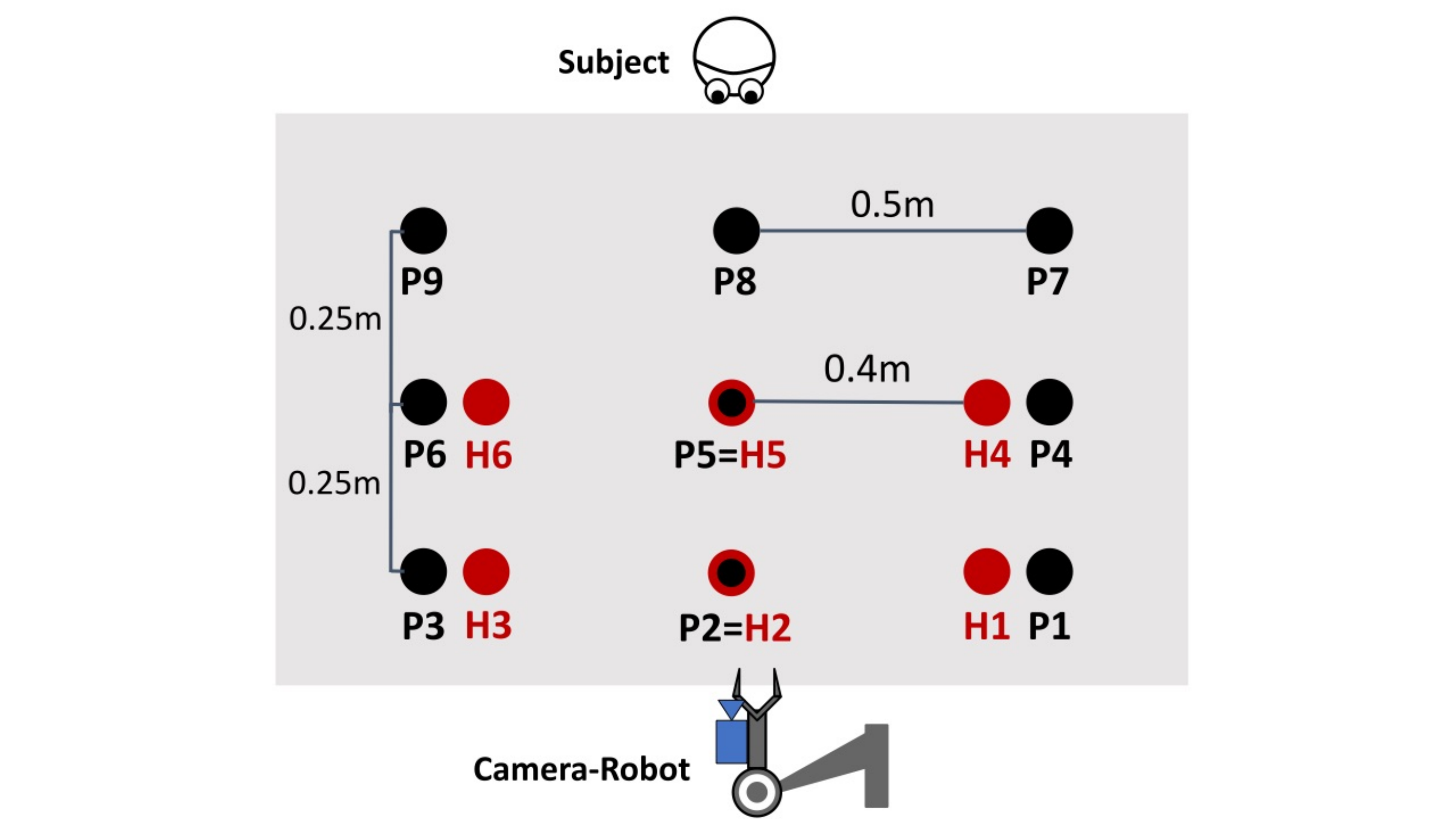}
  \caption{Gaze targets on the shared table, for the object-centered setup. The black and red circles mark targets for head-pose free and head-pose fixed experiments respectively.}
    \label{fig:table_point_grid}
\end{figure}

The gaze targets in the setups are categorized into two groups: \textbf{(i)} The object-centered setup, which features targets distributed across a table representing objects in a shared-workspace. For this, we labeled two sets of visual markers on the shared table (160$\times$70cm) as gaze targets, as depicted in Fig.~\ref{fig:table_point_grid}. The purpose and details of these sets of points will be explained in the related analysis sections. \textbf{(ii)} The mutual-gaze setup, which requires the participant to follow the robot’s camera as a moving target, as shown in Fig. ~\ref{fig:experiment_design_all}\subref{fig:exp-linear}.

\subsection{Participants and Recordings}
\label{sec:participants}
We recruited \textbf{52} participants (\textbf{13} women, \textbf{39} men), aged \textbf{19--57} years (mean \textbf{22.8}, median \textbf{22}). 
A total of \textbf{24} out of participants wore glasses. Our dataset comprises 3,258 videos from \textbf{52 subjects}, totaling \textbf{620,933 frames}.
At 30\,FPS, this corresponds to \textbf{5.7\,hours of video}. All experiment types and the corresponding videos are
balanced per subject, ensuring comparable distributions across participants.

\section{Experiments and Results}
\label{sec:analysis_and_results}

\subsection{Abbreviations and Acronyms}

\textbf{E}: \emph{ETH-X-Gaze} dataset \cite{Zhang2020ETHXGaze},
\textbf{G}: \emph{Gaze360} dataset \cite{kellnhofer2019gaze360}, 
\textbf{SD}: Standard deviation.

\subsection{Experiment 1: Effect of Illumination}
\label{subsec:illumination}
\textit{Research Question 1: Are learning-based gaze estimation methods robust against changes in illumination?}

\begin{figure}[hbt!]
  \centering
  \includegraphics[width=0.75\linewidth]{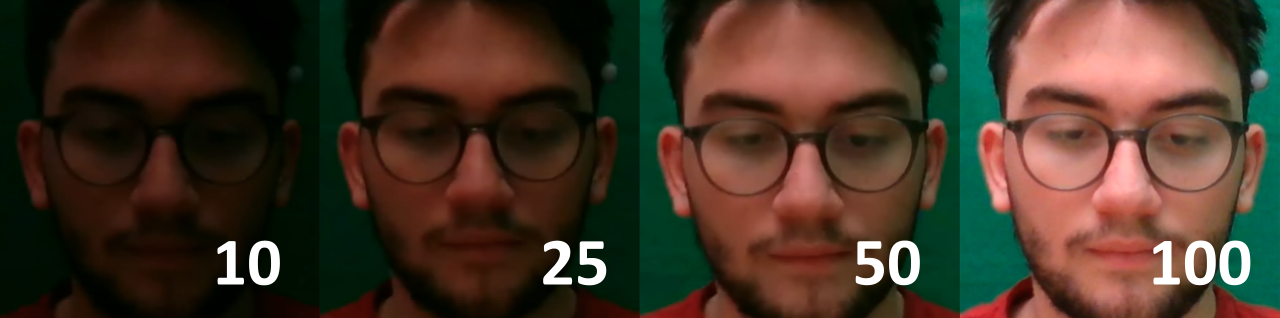}
  \caption{Exp. 1: Illustration of different illumination levels.   }
  \label{fig:gorkem_lighting}
\end{figure}

\noindent\textbf{Setup.}
We use a dimmable light source and experiment with a discrete set of illumination levels: \textit{Illumination 10, 25, 50, 100} (see Fig. \ref{fig:gorkem_lighting} for an example). Each number denotes the dimness level of the light source (e.g., 25 corresponds to 25\% dimness, producing one quarter of the illuminance at 100\% dimness, the maximum supported by the light source). For each level, we recorded gaze for each target point on the grid $\{P_1, \dots, P_9\}$ (as in Fig. \ref{fig:table_point_grid}) such that one video was recorded per point. Head pose is not restricted (the subject is free to naturally orient their head as they gaze at the target point) while the camera (on the robot arm) is fixed and front-facing the subject in this experiment. For each method and level, we compute \emph{subject-level} mean angular error by averaging a subject’s video-level means\footnote{In all experiments ($N=50-52$ subjects, see the Supp. Mat. for details), we use subject-level aggregation to avoid pseudo-replication, as observations for a subject are statistically dependent. Treating the subject as the primary unit ensures our tests satisfy the independence assumptions required for valid statistical inference.}, as shown in \autoref{tab:best-by-lighting}.

\newcommand{\rescell}[2]{\makecell{#1 \\ \footnotesize [#2]}}

\begin{table}[t]
\caption{Exp. 1: Subject-level mean angular error ($^\circ$) and SD. CV (\%) represents the Coefficient of Variation across illumination settings (10, 25, 50, 100). Best methods and their errors are in bold.}
\label{tab:best-by-lighting}
\begin{center}
\scriptsize
\setlength{\tabcolsep}{1.2pt}
\renewcommand{\arraystretch}{1.3}

\begin{tabular}{|>{\raggedright\arraybackslash}p{1.5cm}||c|c|c|c||c|}
\hline
\textbf{Method} & \textbf{10} & \textbf{25} & \textbf{50} & \textbf{100} & \textbf{CV (\%)} \\
\hline
\textbf{PureGaze (E)} & \textbf{11.73 $\pm$ 5.39} & \textbf{11.52 $\pm$ 4.95} & \textbf{12.18 $\pm$ 9.82} & \textbf{10.19 $\pm$ 9.42} & 7.50 \\
\hline
\textbf{GazeTR (E)}   & \textbf{11.50 $\pm$ 3.66} & \textbf{11.16 $\pm$ 3.37} & \textbf{12.45 $\pm$ 7.53} & \textbf{11.50 $\pm$ 7.52} & 4.77 \\
\hline\hline
PureGaze (G) & 16.92 $\pm$ 5.76 & 14.68 $\pm$ 6.23 & 15.13 $\pm$ 6.11 & 17.82 $\pm$ 5.99 & 9.18 \\
\hline
GazeTR (G)   & 14.73 $\pm$ 3.95 & 13.81 $\pm$ 5.16 & 15.27 $\pm$ 5.95 & 17.17 $\pm$ 6.32 & 9.30 \\
\hline
L2CS-Net (G) & 19.61 $\pm$ 8.39 & 17.60 $\pm$ 4.69 & 18.95 $\pm$ 4.56 & 18.99 $\pm$ 3.89 & 4.51 \\
\hline
MCGaze (G)   & 19.08 $\pm$ 6.84 & 14.33 $\pm$ 5.93 & 13.24 $\pm$ 6.26 & 14.29 $\pm$ 5.69 & 17.15 \\
\hline
GaT (G)      & 17.93 $\pm$ 6.01 & 16.74 $\pm$ 6.44 & 15.82 $\pm$ 6.42 & 16.35 $\pm$ 5.89 & 5.36 \\
\hline
\end{tabular}
\end{center}
\end{table}

\noindent\textbf{Results: Method rankings by illumination.} The subject-level mean angular errors are provided in \autoref{tab:best-by-lighting}. Pairwise within-subject $t$-tests (Holm-corrected, $\alpha=0.05$) show that \emph{PureGaze (E)} and \emph{GazeTR (E)} are statistically indistinguishable as the top-performing methods under each illumination setting, consistently outperforming others. Among methods trained on Gaze360, \emph{GazeTR} and \emph{MCGaze} generally share second place; however, their failure modes differ at the extremes: \emph{MCGaze} significantly under-performs in the darkest setting (level 10), while \emph{GazeTR} is significantly more robust in this dark setting. Conversely, at maximum illumination, \emph{MCGaze} emerges as the top-performing Gaze360-trained method, whereas \emph{GazeTR} performance deteriorates substantially in this level. Finally, \emph{GaT} and \emph{L2CS-Net} occupy the bottom rankings.

\noindent\textbf{Results: Robustness to illumination.} Firstly, Tab. \ref{tab:best-by-lighting} provides the Coefficient of Variation ($\textit{CV}$) to summarize how stable each method's accuracy is across different illumination conditions, defined as:
\begin{equation}
\textit{CV} = \sigma / \mu.
\label{eq:cv_calculation}
\end{equation}
where $\sigma$ and $\mu$ denote the standard deviation and mean, respectively, of the method's mean angular errors across the four illumination levels (which are shown in Tab. \ref{tab:best-by-lighting}).
Secondly, training dataset significantly impacts robustness, as methods trained on \emph{ETH-X-Gaze} maintain stable and high accuracy across the entire illumination range. In contrast, methods trained on \emph{Gaze360} appear ``tuned'' to specific illumination conditions, where accuracy degrades as illumination diverges from an optimal level. Specifically, \emph{PureGaze (G)}, \emph{GazeTR (G)}, and \emph{L2CS-Net (G)} exhibit their lowest errors at level 25 illumination, with performance dropping significantly in both darker and brighter settings ($p < .05$). Similarly, \emph{MCGaze (G)} and \emph{GaT (G)} are optimized for 50 illumination, showing a characteristic increase in error as the environment deviates toward darker or brighter. 

\noindent\textbf{Summary and Discussion.}
Regardless of illumination level, \emph{PureGaze (E)} and \emph{GazeTR (E)} deliver the best performance. The \emph{Gaze360}-trained versions of these two architectures, on the other hand, show substantially worse performance. In addition, every Gaze360-trained method seems to be tuned for a particular illumination level (25\% or 50\%), while performing significantly worse at  other levels. This suggests that training on \emph{ETH-X} is extremely helpful for robustness to illumination.

\begin{table}[t]
\caption{Exp. 2: Pairwise tests comparing error in \textit{Fixed Camera} vs. \textit{Camera Viewpoint} setups. Values are subject-level angular error ($^\circ$) (mean $\pm$ SD). $p$-values are two-sided, $p > .05$ (robust methods) are in bold. The best method and its error in \textit{Camera Viewpoint} is in bold, too.}
\label{tab:camera_viewpoint}
\begin{center}
\footnotesize
\setlength{\tabcolsep}{2.4pt}
\renewcommand{\arraystretch}{1.15}
\begin{tabular}{|>{\raggedright\arraybackslash}p{2.00cm}||c|c|c|c|}
\hline
\textbf{Method} & \textbf{Fixed Cam.} & \textbf{Cam. View.} & \textbf{$p$}\\
\hline
\textbf{PureGaze (E)} & 10.19 $\pm$ 9.42 & \textbf{11.12 $\pm$ 4.18} & \textbf{.460} \\
\hline
GazeTR (E)   & 11.50 $\pm$ 7.52 & 14.42 $\pm$ 3.92 & {.004} \\
\hline\hline
PureGaze (G) & 17.82 $\pm$ 5.99 & 18.46 $\pm$ 4.88 & \textbf{.242} \\
\hline
GazeTR (G)   & 17.17 $\pm$ 6.32 & 17.80 $\pm$ 6.44 & \textbf{.308} \\
\hline
L2CS-Net (G) & 18.99 $\pm$ 3.89 & 18.15 $\pm$ 4.32 & {.045} \\
\hline
MCGaze (G)   & 14.29 $\pm$ 5.69 & 15.57 $\pm$ 5.12 & {.034} \\
\hline
GaT (G)      & 16.35 $\pm$ 5.89 & 16.05 $\pm$ 5.88 & \textbf{.486} \\
\hline
\end{tabular}
\end{center}
\end{table}

\subsection{Experiment 2: Effect of Camera Viewpoint}

\textit{Research Question 2: Are learning-based gaze estimation methods robust against camera viewpoint variation?}

\noindent\textbf{Setup.} To analyze the effect of the viewpoint variation, the robot is controlled to move the camera on a circular arc while the subject fixated their gaze on a target point on the table (\autoref{fig:exp-circular}). The robot's motion takes the subject's eye position as the arc's center, its radius as 0.5 meters, and starts the movement at $-30^\circ$ yaw and ends at $+30^\circ$\footnote{These camera yaw values are wrt. to the seating position of the subject.}, spanning $60^\circ$ of viewpoints in total. This movement is repeated for each gaze target point  $\{P_1, \dots, P_9\}$ on the table. Head poses were not restricted and illumination level was set to 100\% dimness, just like \emph{Illumination 100}. In other words, the only difference between \emph{Camera Viewpoint} and the previously discussed \emph{Illumination 100} setting is that there is systematic viewpoint variation within each video in \emph{Camera Viewpoint}, whereas the camera stays in a fixed, $0^\circ$-yaw viewpoint throughout each \emph{Illumination 100} video. Thus, to assess how much each method's performance gets affected by variation in camera viewpoint, we will compare each method's error on the \emph{Camera Viewpoint} to its error on \emph{Illumination 100} (which can be thought of as \emph{Fixed Camera} for this analysis).

\noindent\textbf{Results and Discussion.} Table~\ref{tab:camera_viewpoint} displays the results for this experiment. The analysis highlights that \emph{PureGaze (E)} is the most robust method across varying camera viewpoints, and that \emph{GazeTR (E)} loses its tie with \emph{PureGaze (E)} as the best-performing method.
Moreover, the effect of training dataset is remarkable. Training on \emph{ETH-X} majorly improves performance under viewpoint variation as can be seen from how much better \emph{PureGaze} and \emph{GazeTR} perform when trained on \emph{ETH-X} vs. \emph{Gaze360}.

\subsection{Experiment 3: Effect of Head-Gaze Conflict}

\textit{Research Question 3: Are learning-based gaze estimation methods robust against conflicting head and gaze directions?}

\begin{figure}[tb]
  \centering
  \begin{minipage}[t]{0.35\linewidth}
    \centering
    \includegraphics[width=0.6\linewidth]{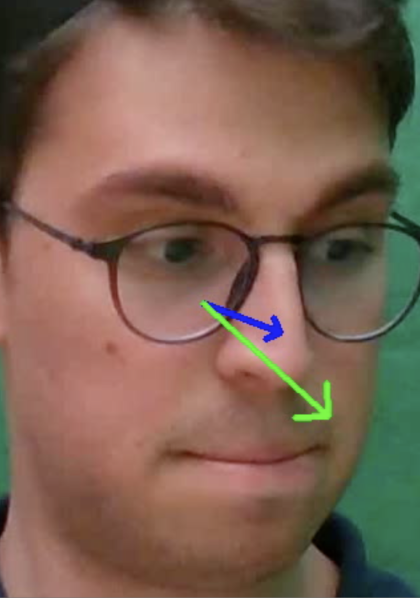}\\
    \textbf{\footnotesize (a) Low Conflict}
  \end{minipage}
  \begin{minipage}[t]{0.35\linewidth}
    \centering
    \includegraphics[width=0.6\linewidth]{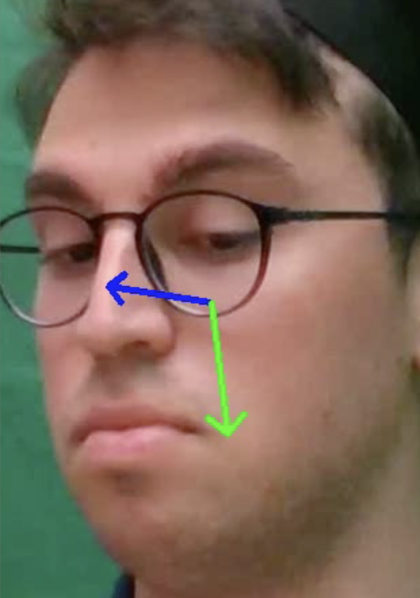}\\
    \textbf{\footnotesize (b) High Conflict}
  \end{minipage}
  \caption{Experiment 3: Samples for low (a) and high (b) levels of head-gaze conflict. Blue arrow: Head direction, Green arrow: Gaze direction.}
  \label{fig:illustrating_head_gaze_conflict}
\end{figure}

\noindent\textbf{Setup.} 
We collected data for three different fixed head orientations: pointing to the left ($+25^\circ$ yaw, $0^\circ$ pitch), center ($0^\circ$ yaw, $0^\circ$ pitch), or right ($-25^\circ$ yaw, $0^\circ$ pitch)
; as described in \autoref{fig:experiment_design_all}\subref{fig:exp-headpose}. For each head pose fixation, gaze was directed at points that would induce varying levels of head-gaze conflict.\footnote{Moreover, this setup's illumination level is set to 100\% dimness and the camera is fixed, similar to the previous \emph{Illumination 100} setting.}

Similar to the definition of \emph{angular error}, we define head-gaze conflict ($\gamma$) as the angular difference between the ground-truth head-forward vector ($\mathbf{h}$) and the ground-truth gaze vector ($\mathbf{g}$):
\begin{equation}
\gamma = \arccos ( (h \cdot g) / (\|h\| \|g\|) ),
\label{eq:head_gaze_conflict}
\end{equation}
To answer ``Does error increase as head-gaze conflict increases?''; for each video $v$ in this setup, we take the mean head--gaze conflict angle $\bar{C}_v$ and the mean angular error $\bar{E}_v$ for the method we test. For each subject $s$, we fit a linear model:
\begin{equation}
\small
\bar{E}_v^{(s)} = \upalpha_{s} + \upbeta_{s}\,\bar{C}_v^{(s)} + \upepsilon_v,
\label{eq:regression_head}
\end{equation}
and test slopes $\{\upbeta_{s}\}$ against zero (one-sample $t$-test, one-sided $H_1\!:\upbeta>0$) with Holm-correction.

\begin{table}[t]
\caption{Experiment 3: Head--gaze conflict results. \emph{Error} shows subject-level angular error ($^\circ$) (mean $\pm$ SD). Following columns display the mean and SD of subject-level slopes $\{\upbeta_s\}$ ($^\circ$ error / $^\circ$ conflict), Holm-corrected $p$ values for the one-sided $t$-test ($H_1:\upbeta_s>0$), and percentage of subjects with positive $\upbeta_s$; respectively. The best method and its error are bolded.}
\label{tab:head_gaze_combined_results}
\begin{center}
\scriptsize
\setlength{\tabcolsep}{3.0pt} \renewcommand{\arraystretch}{1.3}
\begin{tabular}{| l || m{1.45cm} | m{1.45cm} | c | c |}
\hline
\textbf{Method} & \textbf{Error ($^\circ$)} & \textbf{$\upbeta_s$ ($^\circ/^\circ$)} & \textbf{$p\ (H_1:\upbeta_s>0)$} & \textbf{$\%\ \upbeta_s>0$} \\
\hline
\textbf{PureGaze (E)} & \textbf{7.25 $\pm$ 3.65} & +0.05 $\pm$ 0.13 & .023 & 60.0 \\
\hline
GazeTR (E)   & 8.67 $\pm$ 3.71 & -0.05 $\pm$ 0.19 & .950 & 36.0 \\
\hline\hline
PureGaze (G)  & 11.38 $\pm$ 3.32 & +0.04 $\pm$ 0.23 & .248 & 50.0 \\
\hline
GazeTR (G)    & 12.93 $\pm$ 3.74 & +0.38 $\pm$ 0.24 & $2.48 \times 10^{-14}$ & 92.0 \\
\hline
L2CS-Net (G)  & 16.62 $\pm$ 4.77 & +0.50 $\pm$ 0.33 & $4.83 \times 10^{-14}$ & 92.0 \\
\hline
MCGaze (G)    & 20.24 $\pm$ 5.32 & +0.99 $\pm$ 0.30 & $9.39 \times 10^{-28}$ & 100.0 \\
\hline
GaT (G)       & 12.09 $\pm$ 4.99 & +0.20 $\pm$ 0.19 & $4.12 \times 10^{-9}$  & 94.0 \\
\hline
\end{tabular}
\end{center}
\end{table}

\noindent\textbf{Results.} 
Tab. \ref{tab:head_gaze_combined_results} displays results for this experiment. 
\emph{PureGaze (E)} achieves the lowest overall error ($7.25^\circ$), significantly outperforming all other methods ($p_{\textit{Holm}} < .05$ in two-sided pairwise tests between each method's error).

Across subjects and methods, increasing head--gaze conflict is typically associated with higher error, as for most methods $p<.05$ and the percentage of subjects with a positive slope ($\upbeta_s>0$) is $>50\%$. The results reveal that methods trained on \emph{ETH-X} are substantially more robust than those trained on \emph{Gaze360}. Specifically, \emph{GazeTR (E)} is statistically unaffected by conflict, showing a non-significant mean slope ($-0.05^\circ, p_{\textit{Holm}}=.950$). While \emph{PureGaze (E)} has a statistically significant slope ($+0.05^\circ, p_{\textit{Holm}}=.023$), the magnitude of the degradation remains negligible. In contrast, Gaze360-trained methods exhibit much steeper error increases; notably, \emph{MCGaze (G)} shows a slope of $+0.99^\circ$ ($p_{\textit{Holm}} < .001$), indicating that every degree of head-gaze conflict results in a full degree of additional estimation error (in line with the qualitative analysis in \cite{guan2023mcgaze}). Similarly, \emph{L2CS-Net (G)} and \emph{GazeTR (G)} show significant and substantial positive slopes.

\noindent\textbf{Summary and Discussion.} 
\emph{PureGaze (E)} yields the highest zero-shot accuracy in scenarios involving head-gaze conflict. In terms of robustness,
the results indicate the importance of training on the \emph{ETH-X} dataset vs. \emph{Gaze360}, as \emph{ETH-X}-trained methods display superior resilience under head-gaze conflict.

\subsection{Experiment 4: Effect of Gaze Direction}
\textit{Research Question 4: Are learning-based gaze estimation methods robust against variations in gaze direction?}

We investigate this in two setups: The object-centered setup that features gaze targets on a table (comprising \autoref{fig:experiment_design_all}\subref{fig:exp-lighting}, \ \subref{fig:exp-circular},\ \subref{fig:exp-headpose}), and the mutual-gaze setup, where the subject tracks the robot's camera as a moving target (\autoref{fig:exp-linear}). Both setups incorporate an illumination level of 100\% and no head-movement restrictions.

\subsection{Exp. 4.1: Effect of Gaze Dir. in Object-Centered Setup}

\noindent\textbf{Setup.} 
To test the effect of target point location (and thus the corresponding gaze direction), we use the \textit{Illumination 100} setting, which is the most basic setting of the object-centered setup (it lacks challenges such as reduced illumination, head pose restrictions, or camera viewpoint variations discussed in other sections). As discussed in \autoref{subsec:illumination}, this setting comprises 9 different target points in a $3\times3$ grid (shown in \autoref{fig:table_point_grid}) on the table, so corresponding gaze directions are all downward. This is deliberate as downward gaze is particularly important for object-centered interactions; since a person directs their gaze towards the object they are interacting with (like a phone), therefore having a tendency to look downwards \cite{Schreiter2024THORMAGNI}.

\begin{table}[h]
\centering
\caption{Exp. 4.1. Object-Centered Setup: Each cell contains the best method per target point and its error ($^\circ$) (as subject-level mean $\pm$ SD). Darker hues show higher error. }
\label{tab:best-per-point}
\renewcommand{\arraystretch}{1.3}
\setlength{\tabcolsep}{2.5pt}\footnotesize
\begin{tabular}{|c|c|c|c|}
\hline
\multicolumn{4}{|c|}{\textbf{\ \ \ \ \ \ \ \ \ \ \ \ Subject}} \\ \hline
& \textbf{Left} & \textbf{Middle} & \textbf{Right} \\ \hline
\textbf{Closest} & \cellcolor{red!40}\makecell{\textbf{P9:} GazeTR (E) \\ $15.87 \pm 13.07$}& \cellcolor{red!25} \makecell{\textbf{P8:} PureGaze (E) \\ $11.90 \pm 11.46$} & \cellcolor{red!25} \makecell{\textbf{P7:} GazeTR (E) \\ $11.80 \pm 9.18$} \\ \hline
\textbf{Middle} & \cellcolor{red!20} \makecell{\textbf{P6:} PureGaze (E) \\ $10.68 \pm 10.18$} & \cellcolor{red!10} \makecell{\textbf{P5:} PureGaze (E) \\ $7.87 \pm 8.75$} & \cellcolor{red!9}\makecell{\textbf{P4:} PureGaze (E) \\ $7.83 \pm 10.70$} \\ \hline
\textbf{Farthest} & \cellcolor{red!8}\makecell{\textbf{P3:} PureGaze (E) \\ $7.66 \pm 7.60$} & \cellcolor{red!5}\makecell{\textbf{P2:} PureGaze (E) \\ $6.54 \pm 7.93$} & \cellcolor{red!3}\makecell{\textbf{P1:} PureGaze (E) \\ $6.32 \pm 8.78$} \\ \hline
\multicolumn{4}{|c|}{\textbf{\ \ \ \ \ \ \ \ \ \ \ \ Camera}} \\ \hline
\end{tabular}

\end{table}

\noindent\textbf{Results.}
\autoref{tab:best-per-point} presents the best method and the error it achieves for each target point (detailed results are in the Supp. Mat.).\footnote{We use 9 discrete points for statistical analysis rather than the continuous pitch-yaw values to avoid pseudo-replication of gaze directions.} Rows (Closest/Middle/Farthest) are named with respect to the subject, whereas columns (Left/Middle/Right) are from the camera's perspective

Across points, \emph{PureGaze (E)} and \emph{GazeTR (E)} show the best performance, and their performance is statistically indistinguishable ($p_{\text{Holm}}>.05$) except for \emph{P9}.  These two methods consistently outperform \emph{MCGaze}, \emph{GaT}, and \emph{L2CS-Net} across nearly all spatial regions. However, there is substantial variability between points in terms of estimation error, as seen in \autoref{tab:best-per-point}. Thus, we analyze each method's performance based on target position. Since there are too many points, we analyze them by row and by column:

\noindent\textbf{(1) The effect of row-wise distance of the target.}
To formally assess the effect of target distance from the subject, we conduct Holm-corrected pairwise tests between the three target rows and report Cohen effect size ($d$). For the top-performing \emph{PureGaze (E)}, the mean error significantly increases at each step toward the subject ($p_{\textit{Holm}} < .05$). Specifically, error rises from $6.85^\circ$ at the farthest row to $8.78^\circ$ at the middle row ($p_{\textit{Holm}} < .001, |d| = 0.23$), and reaches $14.95^\circ$ at the closest row (middle--closest $p_{\textit{Holm}} < .001, |d| = 0.57$; farthest--closest $p_{\textit{Holm}} < .001, |d| = 0.80$).
\emph{GazeTR (E)} follows a similar trend, too.
The degradation in the closest row is even more pronounced for the Gaze360-trained versions. For \emph{PureGaze (G)}, the error jumps from $14.31^\circ$ at the middle row to $26.91^\circ$ at the closest row ($p_{\textit{Holm}} < .001, |d| = 1.56$), while \emph{GazeTR (G)} rises from $15.13^\circ$ at the middle row to $26.57^\circ$ at the closest row ($p_{\textit{Holm}} < .001, |d| = 1.29$). These results indicate that while close-range downward (steeply downward) gaze is challenging for all methods, training on \emph{ETH-X-Gaze} dataset confers significantly greater spatial resilience.

\noindent\textbf{(2) The effect of left-right orientations (columns).}
\autoref{tab:best-per-point} indicates higher error for points on the left-column. Thus, we conduct subject-level, Holm-corrected pairwise tests between the left and right columns. For \emph{PureGaze (E)}, the left column exhibited significantly higher error than the right column: $12.89^\circ$ vs.\ $8.86^\circ$ ($p_{\textit{Holm}} < .001, |d| = 0.39$). Similarly, \emph{GaT (G)} showed significantly higher error for the left column: $16.46^\circ$ vs.\ $15.17^\circ$ ($p_{\textit{Holm}} = 0.034, |d| = 0.21$). In contrast, \emph{L2CS-Net (G)} exhibited a significant asymmetry in the opposite direction, with the right column showing higher error: $20.15^\circ$ vs.\ $17.74^\circ$ ($p_{\textit{Holm}} = 0.001, |d| = 0.51$). Other methods, including \emph{MCGaze}, showed no significant horizontal asymmetry; \emph{MCGaze} error was $15.31^\circ$ (left) vs.\ $14.10^\circ$ (right) ($p_{\textit{Holm}} = 0.386, |d| = 0.17$), while the Gaze360-trained versions of PureGaze and GazeTR also remained statistically balanced across left/right columns.

\noindent\textbf{Summary and Discussion.} \emph{PureGaze (E)} and \emph{GazeTR} performed best, with \emph{PureGaze} yielding the lowest error for most targets. While most methods were balanced in terms of left-right symmetry, significant asymmetry was observed with \emph{PureGaze (E)}, \emph{GaT (G)}, and \emph{L2CS-Net (G)}. This asymmetry cannot be attributed solely to the training dataset or the architecture; rather, it appears to be the result of their combination.

In addition, spatial depth significantly impacted performance: errors increase for targets closer to the subject. This reflects larger errors for downward gaze, where eyelid occlusion makes gaze estimation naturally difficult.

Performance degrading as gaze becomes more steeply downward has significant implications for HRI, as direct object interaction induces a downward-gaze tendency as people align their gaze with objects of interest \cite{Schreiter2024THORMAGNI,Schreiter2024HeadGaze}. Tasks like using a phone or inspecting a handheld item fall into this \emph{middle/closest} region. Consequently, state-of-the-art methods degrade precisely where fine-grained gaze cues are more useful for interpreting human attention \cite{AdmoniScassellati2017,kompatsiari2017importance,jording2018social}.

\subsection{Exp. 4.2: Effect of Gaze Dir. in Mutual-Gaze Setup}

\noindent\textbf{Setup.}
This experiment was designed to simulate the situation of a person looking into the robot's ``eye''. Hence, the subject maintained their gaze at the robot's camera as the robot-camera moved linearly. The camera moved on horizontal and vertical line segments that make up the \textit{Horizontal Scan} and \textit{Vertical Scan} trajectories (Fig. \ref{fig:experiment_design_all}\subref{fig:exp-linear}). Movement in distinct line segments facilitates analysis of the position of the gaze target, as horizontal movement fixes the vertical position, and vertical movement fixes the horizontal position. Furthermore, unlike the previous object-centered setup, the gaze direction here was centered towards the forward direction of the subject rather than the targets on the table.
In addition, each scan has been conducted with two speeds: \textit{slow} and \textit{fast}, where the motion in \textit{fast} (0.1 m/s) was twice as fast as \textit{slow} (0.05 m/s). Thus, four videos are recorded per subject: \textit{Horizontal Scan-Slow}, \textit{Vertical Scan-Slow}, \textit{Horizontal Scan-Fast}, \textit{Vertical Scan-Fast}.

\noindent\textbf{Results.}
\noindent\textbf{Method performance based on slow/fast and vertical/horizontal scanning conditions.}
We present the subject-level errors in \autoref{tab:tm-summary-merged}. To compare the effect of scan type and movement speed, for each method; we ran Holm-corrected, two sided, subject-level pairwise-tests for each condition-pair from the 4 conditions. Across all methods and condition-pairs, the tests showed that both the scan type and the speed of movement had a negligible effect on error, as the Cohen effect size ($d$) for each comparison was around 0 (even the max $|d|$ was 0.249, where 0.2 is conventionally considered a ``small effect'').

\noindent\textbf{Error vs. Pitch–Yaw Eccentricity.}
To quantify how gaze eccentricity contributes to estimation error, we fit a two-variable regression:
\begin{equation}
\small
E = \upalpha + \upbeta_{\text{pitch}}|\text{Pitch}| + \upbeta_{\text{yaw}}|\text{Yaw}|,
\label{eq:regression_eccentricity}
\end{equation}
where $\upbeta$ coefficients denote the slope of error change ($^\circ$) per absolute pitch or yaw eccentricity ($^\circ$). The results in \autoref{tab:tm-summary-merged} over the 4 conditions highlight the resilience of \emph{PureGaze} and \emph{GazeTR}, and the vulnerability of \emph{MCGaze} to pitch-yaw eccentricity.

\begin{table}[t]
\caption{Exp. 4.2. Mutual-Gaze Setup: Error ($^\circ$) (subject-level mean $\pm$ SD) for each condition (H: Horizontal, V: Vertical). $\textbf{$\upbeta_{\text{pitch}}$}$ \& $\textbf{$\upbeta_{\text{yaw}}$}$: slope of change in error ($^\circ$) per change of absolute pitch \& yaw eccentricity ($^\circ$) (as subject-level mean $\pm$ SD), as aggregated across all four conditions.}
\label{tab:tm-summary-merged}
\begin{center}
\scriptsize
\setlength{\tabcolsep}{2.5pt}
\renewcommand{\arraystretch}{1.8}
\begin{tabular}{|>{\raggedright\arraybackslash}p{1.5cm}||c|c|c|c||c|c|}
\hline
\textbf{Method} & \makecell{\textbf{Fast-H}} & \makecell{\textbf{Fast-V}} & \makecell{\textbf{Slow-H}} & \makecell{\textbf{Slow-V}} & \makecell{\bm{$\upbeta_{\text{pitch}}$}} & \makecell{\bm{$\upbeta_{\text{yaw}}$}} \\
\hline
PureGaze (E) & \makecell{5.38 \\ $\pm$ 2.98} & \makecell{5.31 \\ $\pm$ 2.67} & \makecell{5.41 \\ $\pm$ 3.08} & \makecell{5.30 \\ $\pm$ 3.00} & \makecell{0.08 \\ $\pm$ 0.19} & \makecell{0.01 \\ $\pm$ 0.08} \\
\hline
GazeTR (E)   & \makecell{10.48 \\ $\pm$ 4.29} & \makecell{10.50 \\ $\pm$ 4.23} & \makecell{10.49 \\ $\pm$ 4.72} & \makecell{10.25 \\ $\pm$ 4.45} & \makecell{-0.03 \\ $\pm$ 0.17} & \makecell{0.03 \\ $\pm$ 0.10} \\
\hline\hline
PureGaze (G) & \makecell{9.75 \\ $\pm$ 4.17} & \makecell{9.58 \\ $\pm$ 3.94} & \makecell{9.15 \\ $\pm$ 4.18} & \makecell{9.47 \\ $\pm$ 4.03} & \makecell{0.10 \\ $\pm$ 0.34} & \makecell{-0.01 \\ $\pm$ 0.11} \\
\hline
GazeTR (G)   & \makecell{7.61 \\ $\pm$ 3.46} & \makecell{7.59 \\ $\pm$ 3.12} & \makecell{7.25 \\ $\pm$ 3.84} & \makecell{7.20 \\ $\pm$ 2.29} & \makecell{0.10 \\ $\pm$ 0.23} & \makecell{0.02 \\ $\pm$ 0.06} \\
\hline
L2CS-Net (G) & \makecell{15.33 \\ $\pm$ 2.09} & \makecell{15.83 \\ $\pm$ 2.54} & \makecell{15.23 \\ $\pm$ 2.31} & \makecell{15.41 \\ $\pm$ 2.23} & \makecell{0.30 \\ $\pm$ 0.40} & \makecell{0.66 \\ $\pm$ 0.09} \\
\hline
MCGaze (G)   & \makecell{22.93 \\ $\pm$ 6.99} & \makecell{24.32 \\ $\pm$ 9.18} & \makecell{22.69 \\ $\pm$ 7.53} & \makecell{23.28 \\ $\pm$ 8.74} & \makecell{1.40 \\ $\pm$ 1.01} & \makecell{0.82 \\ $\pm$ 0.29} \\
\hline
GaT (G)      & \makecell{16.43 \\ $\pm$ 2.74} & \makecell{16.89 \\ $\pm$ 3.31} & \makecell{16.26 \\ $\pm$ 2.83} & \makecell{16.23 \\ $\pm$ 2.99} & \makecell{0.49 \\ $\pm$ 0.40} & \makecell{0.65 \\ $\pm$ 0.08} \\
\hline
\end{tabular}
\end{center}
\vspace*{-0.3cm}
\end{table}

\noindent\textbf{Summary and Discussion.} 
Overall, \emph{PureGaze (E)} emerges as the most accurate method in this experiment, too. Regarding the impact of architectural versus dataset-driven factors, the architecture seems to play a more pivotal role than the training dataset for this experiment. Specifically, both the \emph{ETH-X} and \emph{Gaze360}-trained versions of the \emph{PureGaze} and \emph{GazeTR} architectures are significantly more accurate than others. Furthermore, these architectures are much more resilient to gaze eccentricity, with all error/eccentricity slopes $\le 0.10$. While architecture provides a consistent advantage, the effect of the training dataset is mixed: \emph{PureGaze} performs considerably better when trained on \emph{ETH-X} versus \emph{Gaze360}, whereas the opposite is observed for \emph{GazeTR}.

\subsection{Experiment 5: Computational Efficiency}
See the Supp. Mat. for FPS \& memory statistics. 

\section{Conclusion and Discussion}

This study introduces \emph{Gaze4HRI}, a large-scale 3D-gaze benchmark to evaluate appearance-based estimation for HRI. By assessing the state-of-the-art against critical variables like illumination and head-gaze conflict, we reveal these factors as vital differentiators that cause significant degradation across methods. Our findings indicate that the primary requirement for achieving robust performance in appearance-based gaze estimation is a regime that encompasses a large variety of head-pose and gaze direction combinations—while maintaining invariance to subject appearance and illumination—rather than the development of complex models based on spatial-temporal understanding (a recent trend in the literature \cite{guan2023mcgaze,vuillecard2025gat}). Moreover, we observe that the highest accuracy is achieved by using rectified inputs, parallel to prior work \cite{zhang2018revisiting}. In terms of existing resources, the \emph{ETH-X-Gaze} dataset is instrumental, as it provides extensive diversity across 16 illumination conditions and 110 subjects, and most importantly, covers the full spectrum of head-pose and gaze direction combinations. This comprehensive data coverage, when combined with architectural designs such as \emph{PureGaze's} self-adversarial loss to decouple gaze estimation from subject appearance and illumination, provides substantial robustness in the challenging conditions inherent to HRI.

\subsection{Discussion and Key Findings}

\noindent\textbf{Training Dataset.}
Across almost all factors, using \emph{ETH-X-Gaze} as the training dataset over \emph{Gaze360} yielded much higher accuracy and robustness, consistent with observations in the \emph{PureGaze} and \emph{GazeTR} papers~\cite{cheng2022puregaze,cheng2022gazetr}.

\noindent\textbf{Best Method for HRI.} \emph{PureGaze trained on ETH-X-Gaze} uniquely maintains robustness across tested factors (except steeply-downward gaze) and is the most reliable choice for zero-shot estimation.

\noindent\textbf{Open Issues and Guidelines.} Steeply-downward gaze is a universal failure point as eyelid closure is an inherent weakness of appearance-based estimation. An HRI setting with downward gaze may consider using a second camera (preferably below the subject's eye level) other than a robot-mounted camera.

\noindent\textbf{Limitations and Future Work}
Beyond gaze estimation, \emph{Gaze4HRI} facilitates joint estimation by providing synchronous gaze, head pose, and blink annotations in video. While we masked blinks to evaluate current art, real-world systems require robust blink handling; our dataset supports this through video-based temporal cues that make blink detection more robust than single-frame approaches.
\section{ACKNOWLEDGMENTS}
Funded by CoHE/METU BAP (USTA: AI-based Natural Comm. in HRI, No. ADEP-312-2024-11469). We acknowledge METU-ROMER and TÜBİTAK ULAKBİM TRUBA for computational resources. Thanks to B. Akgül, T. Dilsiz, Y. A. Üstün, D. Uzel, E. Kaya, D. Şahin, and S. Karimli.

\section*{ETHICAL IMPACT STATEMENT}
Following the Institutional Review Board approval (0230-ODTUIAEK-2024), facial video data were collected from participants who provided informed consent after being briefed on the study’s nature and intended research use.

\bibliographystyle{ieee}
\bibliography{gaze_estimation}

\clearpage

\appendix

\input{suppl_mat}

\end{document}

%% file: suppl_mat.tex
\def\theequation{S\arabic{equation}}
\renewcommand{\thefigure}{S\arabic{figure}}
\renewcommand{\thesection}{S\arabic{section}}  
\renewcommand{\thetable}{S\arabic{table}}

\subsection{Ground-Truth Validation}
The accuracy of our motion-capture system (OptiTrack) has been validated by tracking the UR5 end-effector simultaneously with OptiTrack and the UR5's internal tracker, which is extremely accurate ($\pm 0.1 \text{ mm}$ \cite{ur5_specs}). Our OptiTrack setup achieved $\pm 0.5 \text{ mm}$ accuracy (aligned with official specs  \cite{optitrack_specs}), confirming its robustness and suitability as ground-truth for this benchmark.

\begin{table*}[t]
\caption{Comparison of Appearance-Based Gaze Datasets (Continued). The table includes the camera resolution, \emph{GT Method} refers to the ground-truth acquisition technique and \emph{Head Restricted} indicates if the subjects' head movement was constrained during data collection, and also key characteristics are summarized.}
\label{tab:dataset_comparison_appendix}
\begin{center}
\footnotesize
\setlength{\tabcolsep}{4pt}
\renewcommand{\arraystretch}{1.3}
\begin{tabular}{|l||c|c|c|m{9cm}|} 
\hline
\textbf{Dataset} & \textbf{Resolution} & \textbf{\makecell{GT\\Method}} & \textbf{\makecell{Head\\Restricted}} & \textbf{Key Characteristics} \\
\hline
\modcell{EYEDIAP}{(2014) \cite{funes2014eyediap}} & VGA/HD & \makecell{3D Floating\\Target} & No & Laboratory dataset featuring 3D floating targets and providing precise 3D ground truth, limited by illumination diversity and a small subject pool. \\
\hline
\modcell{UT Multiview}{(2014) \cite{sugano2014utmultiview}} & 1280x1024 & \makecell{3D Recon.\\(Synthetic)} & Yes & Laboratory dataset using 3D eye reconstruction to synthesize dense training data, provides large-scale coverage of gaze directions; but uses a fixed head pose. \\
\hline
\modcell{MPIIGaze}{(2015) \cite{zhang2017mpiigaze}} & 1280x720 & \makecell{Screen\\Markers} & No & Laptop-based dataset with natural daily variations, limited by narrow gaze/pose range and a small subject pool. \\
\hline
\modcell{GazeCapture}{(2016) \cite{krafka2016gazecapture}} & 640x480 & \makecell{Mobile\\Screen} & No & Large-scale mobile-based dataset featuring extremely high appearance-diversity, but limited range due to screen-camera proximity. \\
\hline
\modcell{RT-GENE}{(2018) \cite{fischer2018rtgene}} & 1920x1080 & \makecell{Eye-tracker\\Glasses} & No & Natural-environment dataset covering longer distances, captured with wearable eye-tracking glasses and thus uses semantic inpainting to remove the hardware from images. \\
\hline
\modcell{Gaze360}{(2019) \cite{kellnhofer2019gaze360}} & 4096x3382 & \makecell{MoCap /\\AlphaPose} & No & In-the-wild dataset with extensive subject and environmental diversity and a full $360^{\circ}$ horizontal range, but has low resolution at far ranges and less-precise ground-truth. \\
\hline
\modcell{ETH-X-Gaze}{(2020) \cite{Zhang2020ETHXGaze}} & 6000x4000 & \makecell{DSLR\\Cluster} & Yes & Laboratory dataset covering all face-visible head poses with very rich head/gaze variability, as well as illumination diversity. \\
\hline
\hline
\modcell{\textbf{Gaze4HRI}}{\textbf{(2025) [Ours]}} & 1920x1080 & \makecell{Robot-\\MoCap} & No & HRI-focused dataset with dynamic robot-camera and target movement, and systematically varied illumination, camera angle, head-gaze poses. \\
\hline
\end{tabular}
\end{center}
\end{table*}

\subsection{Details about the Configuration of Gaze Estimation Methods}

\textbf{PureGaze}~\cite{cheng2022puregaze}: We used the official version pretrained on the \emph{ETH-X-Gaze} dataset \cite{Zhang2020ETHXGaze}. Without making any architectural changes, we also trained a separate version on \emph{Gaze360} \cite{kellnhofer2019gaze360} for 50 epochs, and a constant learning-rate of 0.0001 (based on the official repository). This version scored $10.7457^\circ$ angular error on the test set of \emph{Gaze360}, which is close to the state-of-the-art (which ranges between $10.0-10.5^\circ$)~\cite{cheng2022puregaze, cheng2022gazetr, abdelrahman2022l2cs, guan2023mcgaze, vuillecard2025gat}.

\textbf{GazeTR}~\cite{cheng2022gazetr}: We used the official version pretrained on the \emph{ETH-X-Gaze} dataset. Without making any architectural changes, we also trained a separate version on \emph{Gaze360}~\cite{kellnhofer2019gaze360} for 80 epochs, and a learning-rate of 0.0005 with up warm-up scheduling (based on the official repository). This version achieved an angular error of $11.2336^\circ$ on the \emph{Gaze360} test set. This slight error increase with respect to the state-of-the-art is aligned with the \emph{GazeTR} paper's findings that the architecture should be trained on \emph{ETH-X-Gaze} for optimal performance \cite{cheng2022gazetr}.

\textbf{L2CS-Net}~\cite{abdelrahman2022l2cs}: We used the official weights pretrained on the \emph{Gaze360} dataset.

\textbf{MCGaze}~\cite{guan2023mcgaze}: We used the official release configured with the L2CS-Net backbone and trained on \emph{Gaze360}.

\textbf{GaT}~\cite{vuillecard2025gat}: We evaluated the official implementation trained on \emph{Gaze360}.

\subsection{More Details about Preprocessing (Extracting face region from an image)}

All evaluated methods utilize face regions rather than full image frames. These regions are extracted from the original frames via either: (i) cropping based on a face-detection model, or (ii) data rectification leveraging head-pose information.

\paragraph{Cropping}  
For robust face detection and tracking across video frames, we employed the Ultralytics YOLOv8 face detection method in combination with the built-in BYTETracker algorithm \cite{zhang2022bytetrack,2022yolo5face}. Specifically, we used the pretrained \texttt{yolov8n-face.pt} weights, which provides reliable bounding boxes under real-world conditions. These bounding boxes were then used to generate face crops for input to gaze estimation methods.

\paragraph{Data Rectification} In line with the literature, we applied data rectification as described by Zhang et al. \cite{zhang2018revisiting} and Cheng et al. \cite{cheng2024appearance}\footnote{For data rectification, we directly used the literature-standard code provided at \url{https://phi-ai.buaa.edu.cn/Gazehub/3D-dataset/}} to reduce variability caused by head pose and scale.

\paragraph{Cropping vs Data Rectification} Before deciding which input type (crop vs. rectified) to adopt for each method in the experiments sections of the manuscript, we compared the two configurations for each method across all subjects and frames of the dataset (since each method could be deployed with its best configuration). Table~\ref{tab:crop_vs_rect} reports subject-level angular errors with paired, two-sided $t$-tests (with Holm correction) for the entire \emph{Gaze4HRI} dataset. For \emph{PureGaze} and \emph{GazeTR} architectures; rectification is significantly better, and this holds true whether they are trained on the \emph{ETH-X-Gaze} or \emph{Gaze360} dataset; although the improvement is much greater for the \emph{ETH-X-Gaze} versions.

This superior performance with rectified inputs aligns with prior work demonstrating that data rectification can significantly improve gaze estimation accuracy by canceling out geometric variability \cite{zhang2018revisiting}. However, this is not observed for every method; for \emph{GaT} and \emph{MCGaze} crops clearly outperform rectification\footnote{Guan et al. \cite{guan2023mcgaze} utilize an internal face detection module for MCGaze, which may explain the adverse effect of rectified inputs; notably, the authors do not discuss data rectification. Conversely, Vuillecard et al. \cite{vuillecard2025gat} argue that data rectification is unsuitable for unconstrained settings in GaT, as the process requires reliable head-pose information. Consequently, they do not pursue rectification in their methodology.}, so we use crop versions of these two. For \emph{L2CS-Net}, no statistically significant difference is observed between the two input types. Since cropping is more computationally efficient than rectification---as face detection typically incurs lower overhead than head-pose estimation---and given that this overhead provides no accuracy benefit for \emph{L2CS-Net} (unlike for \emph{PureGaze} and \emph{GazeTR}), we employ the cropped version for \emph{L2CS-Net} throughout our experiments. Accordingly, manuscript experiments use these configurations: \emph{rectified} for \emph{PureGaze} and \emph{GazeTR}, and \emph{cropped} for \emph{L2CS-Net}, \emph{MCGaze}, and \emph{GaT}.

\begin{table}[tb]
\caption{Subject-level mean angular error ($^\circ$) ($\pm$ SD) for crop vs.\ rectified inputs across the full dataset ($N=52$, dof$=51$).}
\label{tab:crop_vs_rect}
\begin{center}
\footnotesize
\renewcommand{\arraystretch}{1.3}
\setlength{\tabcolsep}{4pt}
\begin{tabular}{|l||c|c|c|}
\hline
\textbf{Method} & \textbf{Crop} & \textbf{Rectified} & \textbf{$p_{\text{Holm}}$} \\
\hline
PureGaze (E) & $15.41 \pm 3.57$ & $9.42 \pm 4.21$  & $1.53 \times 10^{-15}$ \\
\hline
GazeTR (E)   & $21.80 \pm 3.69$ & $11.46 \pm 3.54$ & $9.91 \times 10^{-23}$ \\
\hline\hline
PureGaze (G) & $15.53 \pm 3.92$ & $14.42 \pm 4.00$ & .002 \\
\hline
GazeTR (G)   & $19.17 \pm 4.11$ & $13.73 \pm 3.78$ & $5.49 \times 10^{-16}$ \\
\hline
L2CS-Net (G) & $17.58 \pm 3.49$ & $17.08 \pm 4.24$ & .142 \\
\hline
MCGaze (G)   & $17.71 \pm 3.94$ & $57.98 \pm 13.77$ & $1.49 \times 10^{-27}$ \\
\hline
GaT (G)      & $15.69 \pm 4.61$ & $24.51 \pm 5.78$ & $8.55 \times 10^{-27}$ \\
\hline
\end{tabular}
\end{center}
\end{table}

\subsection{Computational Efficiency}
We measured GPU memory usage and inference time on a workstation equipped with an NVIDIA RTX A4000 GPU, and display results in Tab. \ref{tab:efficiency}. For memory, we report the peak allocated VRAM. For inference speed, we report the mean and standard deviation over 1000 runs. The results indicate that single-frame-based architectures \textit{PureGaze}, \textit{GazeTR}, and \textit{L2CS-Net}) are highly efficient whereas multi-frame-based (clip-based) ones (\textit{GaT} and \textit{MCGaze}) are significantly slower. 
The latency incurred by the clip-based architectures may cause higher error in real-time HRI systems, particularly on weaker GPUs.

\begin{table}[!tb]
\caption{Computational efficiency of the methods.\\ Values are mean ($\pm$ SD) ($N=1000$, dof$=999$).}
\label{tab:efficiency}
\begin{center}
\footnotesize
\renewcommand{\arraystretch}{1.2}
\setlength{\tabcolsep}{2pt}
\begin{tabular}{|l||c|c|c|}
\hline
\textbf{Method} & \textbf{Memory (VRAM, MB)} & \textbf{Inference (ms/frame)} & \textbf{FPS} \\
\hline
PureGaze & 142.25  & $1.93 \pm 0.05$ & $518.13$ \\
\hline
GazeTR   & 65.81   & $1.84 \pm 0.04$ & $543.48$ \\
\hline
L2CS-Net & 111.94  & $1.92 \pm 0.14$ & $520.83$ \\
\hline
MCGaze   & 1136.49 & $51.09 \pm 1.66$ & $19.57$ \\
\hline
GaT      & 271.11  & $14.80 \pm 0.11$ & $67.57$ \\
\hline
\end{tabular}
\end{center}
\end{table}

\subsection{Details about Sample Size (Subject Count) For Statistical Inference}
In all experiments, we use subject-level aggregation to avoid pseudo-replication, as observations for a subject are statistically dependent. Treating the subject as the primary unit ensures our tests satisfy the independence assumptions required for valid statistical inference. While most tests utilize a sample size of $N=52$ ($\textit{dof}=51$), minor adjustments were made for data quality. Specifically, Exp. 3 (\emph{head-pose fixed}) excludes two subjects ($N=50, \textit{dof}=49$), and Exp. 4.2 (\emph{mutual-gaze}) excludes one subject ($N=51, \textit{dof}=50$) due to recording errors. These adjustments are noted for transparency and do not materially affect the results.

\subsection{Details about Exp. 3: Effect of Head-Gaze Conflict}
For each head pose fixation, gaze was directed towards the following set of targets: 
Left: $\{H_1, H_2, H_4, H_5\}$, Center: $\{H_1, H_2, H_3, H_4, H_5, H_6\}$, Right: $\{H_2, H_3, H_5, H_6\}$ These sets of targets have been determined such that all three sets comprise points that remain in the field of view of the subject during the head-pose restriction. Fig. 5 of the main paper can be referred to for the layout of target points.

\subsection{Additional Results for Exp. 4: Gaze Direction}

Firstly, Tab. \ref{tab:ground-truth-per-point}
displays the mean pitch \& yaw values ($^\circ$) for the ground-truth gaze vector across subjects under the setting of \emph{Exp. 4.1. Object-Centered Setup}. Rows (Closest, Middle, Farthest) are defined relative to the subject, while columns (Left, Middle, Right) are from the camera's perspective.

Secondly, the following tables provide results for each individual method under this experiment, showing error ($^\circ$) (as subject-level mean $\pm$ SD) for each target point.

\clearpage

\begin{table}[H]
\centering
\caption{Exp. 4.1. Object-Centered Setup: Mean \emph{pitch, yaw} ($^\circ$) of ground-truth gaze vector across subjects for each target point.}
\label{tab:ground-truth-per-point}
\renewcommand{\arraystretch}{1.3}
\setlength{\tabcolsep}{2.5pt}
\begin{tabular}{|c|c|c|c|}
\hline
\multicolumn{4}{|c|}{\textbf{\ \ \ \ \ \ \ \ \ \ \ \ Subject}} \\ \hline
& \textbf{Left} & \textbf{Middle} & \textbf{Right} \\ \hline
\textbf{Closest} & \makecell{\textbf{P9:} Ground Truth \\ $-39.33, 59.39$} & \makecell{\textbf{P8:} Ground Truth \\ $-60.29, 4.26$} & \makecell{\textbf{P7:} Ground Truth \\ $-40.25, -58.69$} \\ \hline
\textbf{Middle} & \makecell{\textbf{P6:} Ground Truth \\ $-32.61, 42.99$} & \makecell{\textbf{P5:} Ground Truth \\ $-40.69, 0.81$} & \makecell{\textbf{P4:} Ground Truth \\ $-31.82, -42.05$} \\ \hline
\textbf{Farthest} & \makecell{\textbf{P3:} Ground Truth \\ $-26.11, 32.24$} & \makecell{\textbf{P2:} Ground Truth \\ $-29.48, 0.01$} & \makecell{\textbf{P1:} Ground Truth \\ $-24.90, -32.35$} \\ \hline
\multicolumn{4}{|c|}{\textbf{\ \ \ \ \ \ \ \ \ \ \ \ Camera}} \\ \hline
\end{tabular}
\end{table}

\begin{table}[H]
\centering
\caption{Exp. 4.1. Object-Centered Setup: Error ($^\circ$) (as subject-level mean $\pm$ SD) of \emph{PureGaze (E)} for each target point.}
\label{tab:puregaze-e-per-point}
\renewcommand{\arraystretch}{1.3}
\setlength{\tabcolsep}{2.5pt}
\begin{tabular}{|c|c|c|c|}
\hline
\multicolumn{4}{|c|}{\textbf{\ \ \ \ \ \ \ \ \ \ \ \ Subject}} \\ \hline
& \textbf{Left} & \textbf{Middle} & \textbf{Right} \\ \hline
\textbf{Closest} & \makecell{\textbf{P9:} PureGaze (E) \\ $20.41 \pm 14.91$} & \makecell{\textbf{P8:} PureGaze (E) \\ $11.90 \pm 11.46$} & \makecell{\textbf{P7:} PureGaze (E) \\ $12.48 \pm 13.85$} \\ \hline
\textbf{Middle} & \makecell{\textbf{P6:} PureGaze (E) \\ $10.68 \pm 10.18$} & \makecell{\textbf{P5:} PureGaze (E) \\ $7.87 \pm 8.75$} & \makecell{\textbf{P4:} PureGaze (E) \\ $7.83 \pm 10.70$} \\ \hline
\textbf{Farthest} & \makecell{\textbf{P3:} PureGaze (E) \\ $7.66 \pm 7.60$} & \makecell{\textbf{P2:} PureGaze (E) \\ $6.54 \pm 7.93$} & \makecell{\textbf{P1:} PureGaze (E) \\ $6.32 \pm 8.78$} \\ \hline
\multicolumn{4}{|c|}{\textbf{\ \ \ \ \ \ \ \ \ \ \ \ Camera}} \\ \hline
\end{tabular}
\end{table}

\begin{table}[H]
\centering
\caption{Exp. 4.1. Object-Centered Setup: Error ($^\circ$) (as subject-level mean $\pm$ SD) of \emph{GazeTR (E)} for each target point.}
\label{tab:gazetr-e-per-point}
\renewcommand{\arraystretch}{1.3}
\setlength{\tabcolsep}{2.5pt}
\begin{tabular}{|c|c|c|c|}
\hline
\multicolumn{4}{|c|}{\textbf{\ \ \ \ \ \ \ \ \ \ \ \ Subject}} \\ \hline
& \textbf{Left} & \textbf{Middle} & \textbf{Right} \\ \hline
\textbf{Closest} & \makecell{\textbf{P9:} GazeTR (E) \\ $15.87 \pm 13.07$} & \makecell{\textbf{P8:} GazeTR (E) \\ $14.04 \pm 9.47$} & \makecell{\textbf{P7:} GazeTR (E) \\ $11.80 \pm 9.18$} \\ \hline
\textbf{Middle} & \makecell{\textbf{P6:} GazeTR (E) \\ $11.46 \pm 10.77$} & \makecell{\textbf{P5:} GazeTR (E) \\ $11.23 \pm 7.23$} & \makecell{\textbf{P4:} GazeTR (E) \\ $9.47 \pm 5.96$} \\ \hline
\textbf{Farthest} & \makecell{\textbf{P3:} GazeTR (E) \\ $10.12 \pm 9.80$} & \makecell{\textbf{P2:} GazeTR (E) \\ $9.64 \pm 6.98$} & \makecell{\textbf{P1:} GazeTR (E) \\ $10.11 \pm 5.36$} \\ \hline
\multicolumn{4}{|c|}{\textbf{\ \ \ \ \ \ \ \ \ \ \ \ Camera}} \\ \hline
\end{tabular}
\end{table}

\begin{table}[H]
\centering
\caption{Exp. 4.1. Object-Centered Setup: Error ($^\circ$) (as subject-level mean $\pm$ SD) of \emph{PureGaze (G)} for each target point.}
\label{tab:puregaze-g-per-point}
\renewcommand{\arraystretch}{1.3}
\setlength{\tabcolsep}{2.5pt}
\begin{tabular}{|c|c|c|c|}
\hline
\multicolumn{4}{|c|}{\textbf{\ \ \ \ \ \ \ \ \ \ \ \ Subject}} \\ \hline
& \textbf{Left} & \textbf{Middle} & \textbf{Right} \\ \hline
\textbf{Closest} & \makecell{\textbf{P9:} PureGaze (G) \\ $30.36 \pm 15.61$} & \makecell{\textbf{P8:} PureGaze (G) \\ $19.21 \pm 8.85$} & \makecell{\textbf{P7:} PureGaze (G) \\ $31.30 \pm 11.95$} \\ \hline
\textbf{Middle} & \makecell{\textbf{P6:} PureGaze (G) \\ $13.75 \pm 7.21$} & \makecell{\textbf{P5:} PureGaze (G) \\ $12.53 \pm 7.87$} & \makecell{\textbf{P4:} PureGaze (G) \\ $16.67 \pm 9.52$} \\ \hline
\textbf{Farthest} & \makecell{\textbf{P3:} PureGaze (G) \\ $13.00 \pm 4.59$} & \makecell{\textbf{P2:} PureGaze (G) \\ $11.63 \pm 6.68$} & \makecell{\textbf{P1:} PureGaze (G) \\ $11.98 \pm 5.90$} \\ \hline
\multicolumn{4}{|c|}{\textbf{\ \ \ \ \ \ \ \ \ \ \ \ Camera}} \\ \hline
\end{tabular}
\end{table}

\begin{table}[H]
\centering
\caption{Exp. 4.1. Object-Centered Setup: Error ($^\circ$) (as subject-level mean $\pm$ SD) of \emph{GazeTR (G)} for each target point.}
\label{tab:gazetr-g-per-point}
\renewcommand{\arraystretch}{1.3}
\setlength{\tabcolsep}{2.5pt}
\begin{tabular}{|c|c|c|c|}
\hline
\multicolumn{4}{|c|}{\textbf{\ \ \ \ \ \ \ \ \ \ \ \ Subject}} \\ \hline
& \textbf{Left} & \textbf{Middle} & \textbf{Right} \\ \hline
\textbf{Closest} & \makecell{\textbf{P9:} GazeTR (G) \\ $30.06 \pm 19.34$} & \makecell{\textbf{P8:} GazeTR (G) \\ $19.13 \pm 11.89$} & \makecell{\textbf{P7:} GazeTR (G) \\ $30.47 \pm 11.00$} \\ \hline
\textbf{Middle} & \makecell{\textbf{P6:} GazeTR (G) \\ $16.17 \pm 7.68$} & \makecell{\textbf{P5:} GazeTR (G) \\ $12.28 \pm 5.94$} & \makecell{\textbf{P4:} GazeTR (G) \\ $16.94 \pm 6.74$} \\ \hline
\textbf{Farthest} & \makecell{\textbf{P3:} GazeTR (G) \\ $10.92 \pm 5.76$} & \makecell{\textbf{P2:} GazeTR (G) \\ $7.86 \pm 5.09$} & \makecell{\textbf{P1:} GazeTR (G) \\ $10.62 \pm 4.17$} \\ \hline
\multicolumn{4}{|c|}{\textbf{\ \ \ \ \ \ \ \ \ \ \ \ Camera}} \\ \hline
\end{tabular}
\end{table}

\begin{table}[H]
\centering
\caption{Exp. 4.1. Object-Centered Setup: Error ($^\circ$) (as subject-level mean $\pm$ SD) of \emph{L2CS-Net (G)} for each target point.}
\label{tab:l2cs-net-g-per-point}
\renewcommand{\arraystretch}{1.3}
\setlength{\tabcolsep}{2.5pt}
\begin{tabular}{|c|c|c|c|}
\hline
\multicolumn{4}{|c|}{\textbf{\ \ \ \ \ \ \ \ \ \ \ \ Subject}} \\ \hline
& \textbf{Left} & \textbf{Middle} & \textbf{Right} \\ \hline
\textbf{Closest} & \makecell{\textbf{P9:} L2CS-Net (G) \\ $24.83 \pm 7.82$} & \makecell{\textbf{P8:} L2CS-Net (G) \\ $29.68 \pm 12.30$} & \makecell{\textbf{P7:} L2CS-Net (G) \\ $28.86 \pm 8.01$} \\ \hline
\textbf{Middle} & \makecell{\textbf{P6:} L2CS-Net (G) \\ $16.81 \pm 4.20$} & \makecell{\textbf{P5:} L2CS-Net (G) \\ $16.46 \pm 5.12$} & \makecell{\textbf{P4:} L2CS-Net (G) \\ $18.75 \pm 4.87$} \\ \hline
\textbf{Farthest} & \makecell{\textbf{P3:} L2CS-Net (G) \\ $11.64 \pm 3.86$} & \makecell{\textbf{P2:} L2CS-Net (G) \\ $10.94 \pm 4.53$} & \makecell{\textbf{P1:} L2CS-Net (G) \\ $12.81 \pm 4.43$} \\ \hline
\multicolumn{4}{|c|}{\textbf{\ \ \ \ \ \ \ \ \ \ \ \ Camera}} \\ \hline
\end{tabular}
\end{table}

\begin{table}[H]
\centering
\caption{Exp. 4.1. Object-Centered Setup: Error ($^\circ$) (as subject-level mean $\pm$ SD) of \emph{MCGaze (G)} for each target point.}
\label{tab:mcgaze-g-per-point}
\renewcommand{\arraystretch}{1.3}
\setlength{\tabcolsep}{2.5pt}
\begin{tabular}{|c|c|c|c|}
\hline
\multicolumn{4}{|c|}{\textbf{\ \ \ \ \ \ \ \ \ \ \ \ Subject}} \\ \hline
& \textbf{Left} & \textbf{Middle} & \textbf{Right} \\ \hline
\textbf{Closest} & \makecell{\textbf{P9:} MCGaze (G) \\ $16.75 \pm 10.32$} & \makecell{\textbf{P8:} MCGaze (G) \\ $14.02 \pm 8.42$} & \makecell{\textbf{P7:} MCGaze (G) \\ $16.35 \pm 10.04$} \\ \hline
\textbf{Middle} & \makecell{\textbf{P6:} MCGaze (G) \\ $15.50 \pm 9.13$} & \makecell{\textbf{P5:} MCGaze (G) \\ $15.65 \pm 8.86$} & \makecell{\textbf{P4:} MCGaze (G) \\ $13.90 \pm 7.44$} \\ \hline
\textbf{Farthest} & \makecell{\textbf{P3:} MCGaze (G) \\ $13.62 \pm 6.73$} & \makecell{\textbf{P2:} MCGaze (G) \\ $10.54 \pm 7.12$} & \makecell{\textbf{P1:} MCGaze (G) \\ $11.97 \pm 6.91$} \\ \hline
\multicolumn{4}{|c|}{\textbf{\ \ \ \ \ \ \ \ \ \ \ \ Camera}} \\ \hline
\end{tabular}
\end{table}

\begin{table}[H]
\centering
\caption{Exp. 4.1. Object-Centered Setup: Error ($^\circ$) (as subject-level mean $\pm$ SD) of \emph{GaT (G)} for each target point.}
\label{tab:gat-g-per-point}
\renewcommand{\arraystretch}{1.3}
\setlength{\tabcolsep}{2.5pt}
\begin{tabular}{|c|c|c|c|}
\hline
\multicolumn{4}{|c|}{\textbf{\ \ \ \ \ \ \ \ \ \ \ \ Subject}} \\ \hline
& \textbf{Left} & \textbf{Middle} & \textbf{Right} \\ \hline
\textbf{Closest} & \makecell{\textbf{P9:} GaT (G) \\ $20.84 \pm 8.96$} & \makecell{\textbf{P8:} GaT (G) \\ $21.54 \pm 9.96$} & \makecell{\textbf{P7:} GaT (G) \\ $20.49 \pm 7.85$} \\ \hline
\textbf{Middle} & \makecell{\textbf{P6:} GaT (G) \\ $15.76 \pm 6.99$} & \makecell{\textbf{P5:} GaT (G) \\ $17.25 \pm 6.90$} & \makecell{\textbf{P4:} GaT (G) \\ $13.70 \pm 6.09$} \\ \hline
\textbf{Farthest} & \makecell{\textbf{P3:} GaT (G) \\ $12.76 \pm 6.03$} & \makecell{\textbf{P2:} GaT (G) \\ $13.20 \pm 6.52$} & \makecell{\textbf{P1:} GaT (G) \\ $11.35 \pm 5.77$} \\ \hline
\multicolumn{4}{|c|}{\textbf{\ \ \ \ \ \ \ \ \ \ \ \ Camera}} \\ \hline
\end{tabular}
\end{table}